%% file: Cogradient.tex
\documentclass{article}

\pdfoutput=1
\usepackage{jmlr2e}
\usepackage{times}
\usepackage{epsfig}
\usepackage{graphicx}
\usepackage{amsmath}
\usepackage{amssymb}

\usepackage{graphicx}
\usepackage{float}
\usepackage{subfig}
\usepackage{dsfont}
\usepackage{algorithm}
\usepackage{algorithmic}
\usepackage{bm}
\usepackage{booktabs}
\usepackage{multirow}
\graphicspath{{figures/}}
\firstpageno{1}

\begin{document}

\title{Cogradient Descent for Dependable Learning}

\author{\name Runqi Wang \email runqiwang@buaa.edu.cn \\
       \addr School of Automation Science and Electrical Engineering\\
       Beihang University\\
       Beijing, 100089, China
       \AND
       \name Baochang Zhang* \email         bczhang@139.com \\
       \addr School of Automation Science and Electrical Engineering\\
        Beihang University\\
       Beijing, 100089, China
       \AND
       \name Li'an Zhuo \email lianzhuo@buaa.edu.cn \\
       \addr School of Automation Science and Electrical Engineering\\
       Beihang University\\
       Beijing, 100089, China
       \AND
       \name Qixiang Ye \email qxye@ucas.ac.cn \\
       \addr School of Electronic.Electrical and Communication Engineering\\
       University of Chinese Academy and Sciences\\
       Beijing, 100089, China
       \AND
       \name David Doermann \email Doermann@buffalo.edu \\
       \addr University at Buffalo\\
       Buffalo, 14200, USA}

\editor{}

\maketitle

\begin{abstract}
Conventional gradient descent methods compute the gradients for multiple variables through the partial derivative. Treating the coupled variables independently while ignoring the interaction, however, leads to an insufficient optimization for bilinear models. In this paper, we propose a dependable learning based on Cogradient Descent (CoGD) algorithm to address the bilinear optimization problem, providing a systematic way to coordinate the gradients of coupling variables based on a kernelized projection function. CoGD is introduced to solve  bilinear problems when one variable is with  sparsity constraint, as often occurs in modern learning paradigms. CoGD can also be used to decompose the association of features and weights, which further generalizes our method to better train convolutional neural networks (CNNs) and improve the model capacity. CoGD is applied in representative bilinear problems, including image reconstruction, image inpainting, network pruning and CNN training. Extensive experiments show that CoGD improves the state-of-the-arts by significant margins. Code is available at {https://github.com/bczhangbczhang/CoGD}.
\end{abstract}


\section{Introduction}

Gadient descent prevails in performing optimization in computer vision and machine learning. As one of its widest uses, back propagation (BP) exploits the gradient descent algorithm to learn model (neural network) parameters by pursuing the minimum value of a loss function. Previous studies have focused on improving the gradient descent algorithm to make the loss decrease faster and more stably~\cite{Kingma2014Adam,Dozat2016Incorporating,Zeiler2012ADADELTA}. However, most of them use the classical partial derivative to calculate the gradients without considering the intrinsic relationship between variables, especially in terms of the convergence speed. We observed in many learning applications, that the variable of the sparsity constraint converges faster than the variables without constraint~\cite{CoGD2020}. This implies that the variables are coupled in terms of convergence speed, and such a coupling relationship should be considered in the optimization, yet  remains largely unexplored.
		
	    Bilinear optimization models are the cornerstone of many computer vision algorithms. Often, the optimized objectives or models are influenced by two or more hidden factors that interact to produce the observations~\cite{heide2015fast,yang2017image} . With bilinear models, we can disentangle the variables, $e.g.$, the illumination and object colors in color constancy, the shape from shading, or object identity and its pose in recognition. Such models have shown great potential in extensive low-level applications, including deblurring \cite{Young2019solving}, denoising \cite{abdelhamed2019noise}, and $3$D object reconstruction \cite{del2011bilinear}. They have also evolved in convolutional neural networks (CNNs), leading to promising applications with model and feature interactions that are particularly useful for fine-grained categorization and model pruning \cite{liao2019squeezed,liu2017Learning}. 
	
    	A basic bilinear optimization problem~\cite{mairal2010online}  attempts to optimize the following objective function as
    	\begin{equation}
    	\label{eq:bl_sparse}
    	\mathop {\arg \min }\limits_{\mathbf{A},\mathbf{x}} G(\mathbf{A},\mathbf{x}) = \|\mathbf{b} - \mathbf{A}\mathbf{x}\|_2^2\\+\lambda\|\mathbf{x}\|_1+R(A),
	    \end{equation}
    	where $\mathbf{b} \in \mathbb{R}^{M \times 1}$ is an observation that can be characterized by $\mathbf{A}\in \mathbb{R}^{M \times N}$ and $\mathbf{x} \in \mathbb{R}^{N \times 1}$. $R(\cdot)$ represents the regularization, typically the $\ell_1$ or $\ell_2$ norm. $\|\mathbf{b} - \mathbf{A}\mathbf{x}\|_2^2$ can be replaced by any function with the form $\mathbf{A}\mathbf{x}$. Bilinear models generally have one variable with a sparsity constraint such as $\ell_1$ regularization with the aim to avoid overfitting.

	    \begin{figure}[!t]
		\centering
		\includegraphics[width=0.95\linewidth]{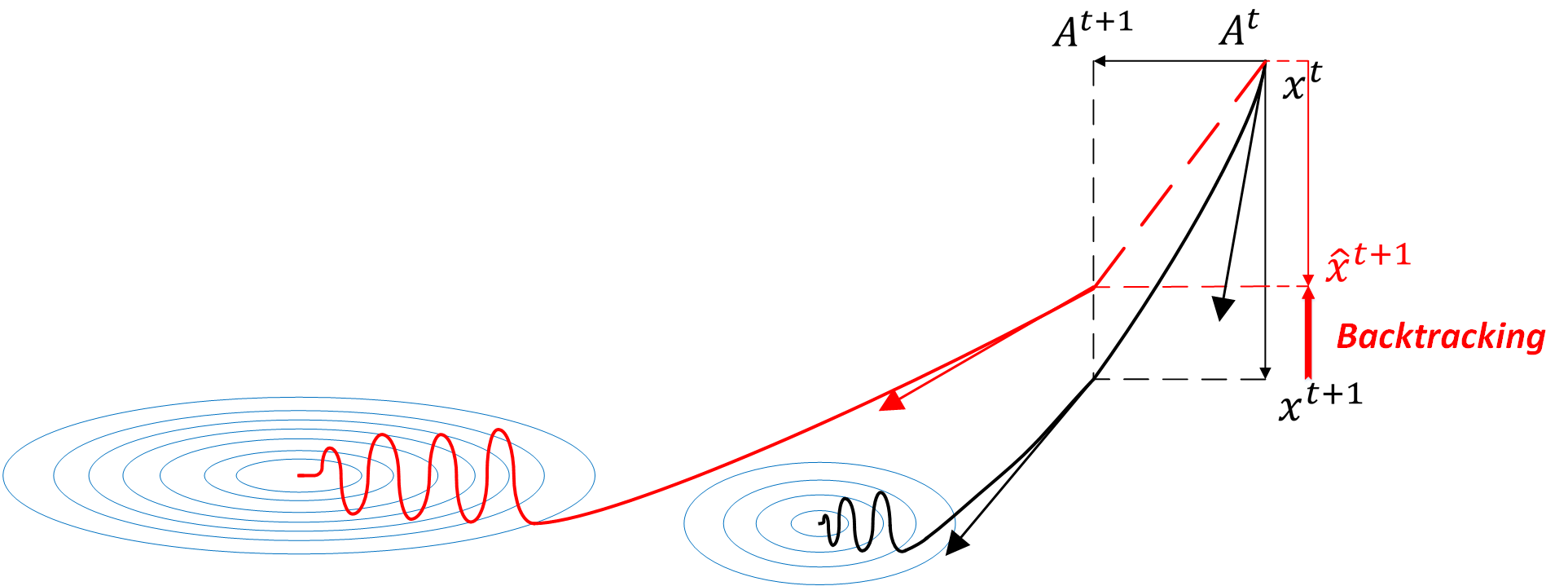}%
		\caption{CoGradient Descent (CoGD) drives the optimization escaping from local minima by backtracking (projecting) the current solution to its previous positions based on a projection function.} 
		\label{csc1}
	    \end{figure}

		Existing methods tend to decompose the bilinear optimization problem into manageable sub-problems, and solve them using the Alternating Direction Method of Multipliers (ADMM) ~\cite{heide2015fast,yang2017image} . Without considering the relationship between two hidden factors, however, existing methods suffer from sub-optimal solutions caused by an asynchronous convergence speed of the hidden variables. The variable with the sparsity constraint can converge faster than the variables without the sparsity constraint, which causes insufficient training since the optimization with less active variables is more likely to fall into the local minima. In terms of the convergence speed, the interaction between different variables is seldom explored in the literature so that the collaborative or dependable  nature of different variables for optimization is neglected, leading to a sub-optimal solution.  As shown in Fig.~\ref{csc1}, we backtrack the sparse variable that will be collaborating again with other variables to facilitate the escape from local minima and pursuing optimal solutions for complex tasks.
	
	    In this paper, we introduce a Cogradient Descent algorithm (CoGD) for dependable learning on bilinear models, and target the asynchronous gradient descent problem by considering its coupled relationship with other variables. CoGD is formulated as a general framework where the coupling between hidden variables can be used to coordinate the gradients based on a kernelized projection function. Based on CoGD, the variables are sufficiently trained and decoupled to improve the training process.  As shown in \cite{causality}, the resulting decoupling variables can enhance the causality of the learning system.    CoGD is applied to representative bilinear problems with one variable having a sparsity constraint, which is widely used in the learning paradigm. The contributions of this work are summarized as follows:

    	\begin{itemize}
		\item
		We propose a dependable learning based on Cogradient Descent (CoGD) algorithm to better solve the bilinear optimization, creating a solid theoretical framework to coordinate the gradient of hidden variables based on a kernelized projection function. 
		
		\item
		We propose an optimization strategy that considers the interaction of variables in bilinear optimization, and solve the asynchronous convergence  with gradient-based learning procedures.
		
		\item
		Extensive experiments demonstrate that CoGD achieves significant performance improvements on typical bilinear problems including convolutional sparse coding (CSC), network pruning, and CNN training. 
	    \end{itemize}
	
    	This work is an extension of our CVPR paper~\cite{CoGD2020} by providing the details of the derivation of our theoretical model, which leverages a kernelized projection function to reveal the interaction of the variables in terms of convergence speed. In addition, more extensive experiments including CNN model training are conducted to validate the effectiveness of the proposed algorithm. 
	
        \section{Related Work}\label{sec:relatedwork}

        \textbf{Gradient Descent.} Gradient descent plays one of the essential roles in the optimization of differentiable models, by pursuing a solution for an objective function to minimize the cost function, as far as possible. It starts from an initial state that is iteratively updated by the opposite partial derivative with respect to the current input. Variations in the gradient update method lead to different versions of gradient descent, such as Momentum, Adaptive Moment Estimation~\cite{Kingma2014Adam}, Nesterov accelerated gradient~\cite{Dozat2016Incorporating}, and Adagrad~\cite{Zeiler2012ADADELTA}.

        In the deep learning era, with large-scale dataset, stochastic gradient descent (SGD) and its variants are practical choices. Unlike vanilla gradient descent which performs the update based on the entire training set, SGD can work well using solely a small part of the training data. SGD is generally termed batch SGD and allows the optimizer converge to minima or local minima based on gradient descent. There are some related literature~\cite{goldt2019dynamics} which discuss the dynamic properties of SGD for deep neural networks. These dynamics and their performance are investigated in the teacher-student setup using SGD, which shows how the dynamics of SGD are captured by a set of differential equations. It also indicates that achieving good generalization in neural networks goes beyond the properties of SGD alone and depends on the interplay of the algorithm, the model architecture, and the data distribution.
	
	    \textbf{Convolutional Sparse Coding.} Convolutional sparse coding (CSC) is a classic bilinear optimization problem and has been exploited for image reconstruction and inpainting. Existing algorithms for CSC tend to split the bilinear optimization problem into subproblems, each of which is iteratively solved by ADMM. Fourier domain approaches~\cite{bristow2013fast,wohlberg2014efficient} are also exploited to solve the $\ell_1$ regularization sub-problem using soft thresholding. Furthermore, recent works ~\cite{heide2015fast,yang2017image} split the objective into a sum of convex functions and introduce ADMM with proximal operators to speed up the convergence. Although the generalized bilinear model~\cite{yokoya2012generalized} considers a nonlinear combination of several end members in one matrix (not two bilinear matrices), it only proves to be effective in unmixing hyperspectral images. These approaches simplify bilinear optimization problems by regarding the two factors as independent and optimizing one variable while keeping the other unchanged.
	    
	    \textbf{Bilinear Models in Deep Learning.} Bilinear models can be embedded in CNNs. One application is network pruning, which is one of the hottest topics in the deep learning community~\cite{Lin2019Towards, lin2020hrank, liu2017Learning}. With the aid of bilinear models, the important feature maps and corresponding channels are pruned ~\cite{liu2017Learning}. Bilinear based network pruning can be performed by iterative methods like modified the Accelerated Proximal Gradient (APG)~\cite{Huang2017Data} and the iterative shrinkage-thresholding algorithm (ISTA)~\cite{ye2018rethinking,Lin2019Towards}.  A number of deep learning applications, such as  fine-grained categorization~\cite{lin2015bilinear,li2017factorized}, visual question answering (VQA)~\cite{yu2017multi} and person re-identification~\cite{suh2018part},  attempt to embed bilinear models into CNNs to model pairwise feature interactions and fuse multiple features with attention. To update the parameters, they directly utilize the gradient descent algorithm and back-propagate the gradients of the loss.
	
    	An interesting  application of  CoGD is studied in CNNs learning. Considering the linearity of the convolutional operation, CNN training can also be considered as a bilinear optimization task   as 
	    \begin{equation}
	    F_j^{l + 1} =  f(BN(\sum\limits_i {F_i^l \otimes W_{{{i,j}}}^l})),
	    \label{eq_convbc}
	    \end{equation}
	    where \(F_j^{l}\) and \(F_j^{l + 1}\) are the \( i \)-th input  and the \( j \)-th output feature maps at the $l$-th and $(l+1)$-th layer, \( W^l_{i,j}\) are convolutional filters, and \( \otimes \), \( BN( \cdot )\) and \( f( \cdot )\) refer to convolutional operator, batch normalization and activation respectively. However, the convolutional operation is not as efficient as a traditional bilinear model.  We instead consider a batch normalization (BN) layer to validate our method, which can be formulated as a bilinear optimization problem as detailed in Section 4.2. We  use the CoGD to replace SGD to efficiently learn the CNN, with the aim to validate the effectiveness of the proposed method.
	   
	    \begin{figure*}[!t]
		\centering
		\includegraphics[width=1\linewidth]{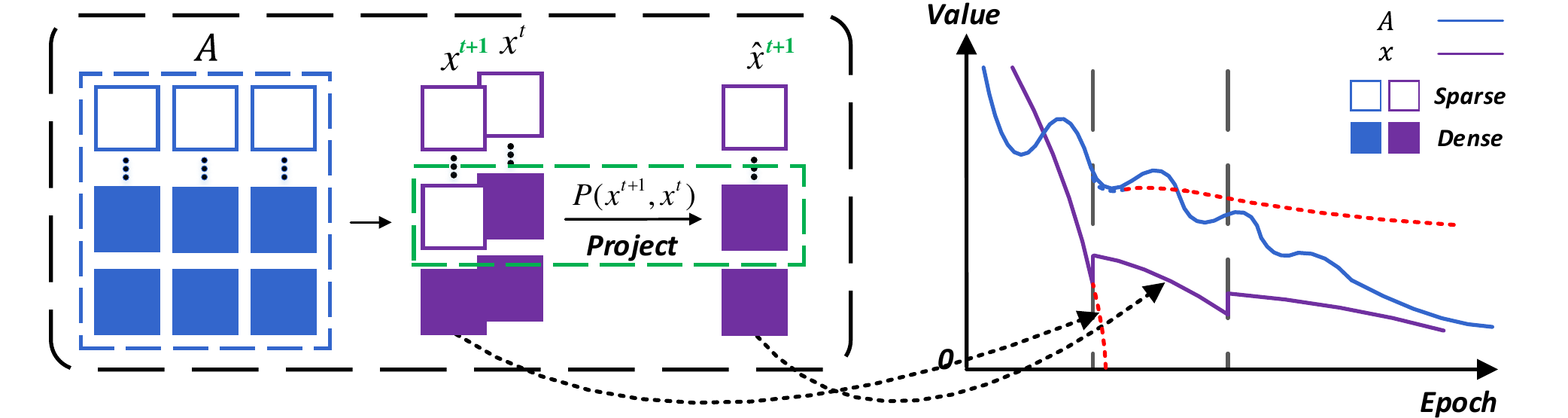}
		\caption{An illustration of CoGD. Conventional gradient-based algorithms have a heuristic that the two hidden variables in bilinear models are independent. Nevertheless, we validate that such a heuristic is implausible, and the algorithms suffer from asynchronous convergence and sub-optimal solutions. The red dotted lines denote that the sparsity of $\mathbf{x}$ causes an inefficient training of $\mathbf{A}$. We propose a projection function to coordinate the gradient of the hidden variables.} 
		\label{csc1}
	    \end{figure*}

\section{The Proposed Method}
	The proposed method considers the relationship between two variables with the benefits on the linear inference. It is essentially different from previous bilinear models which typically optimize one variable while keeping the other unchanged. In what follows, we first discuss the gradient varnishing problem in bilinear models and then present CoGD.
	
	\subsection{Gradient Descent}
	
	\noindent Assuming $\mathbf{A}$ and $\mathbf{x}$ are independent, the conventional gradient descent method can be used to solve the bilinear optimization problem as
	\begin{equation}
	\mathbf{A}^{t+1} = \mathbf{A}^{t} + \eta_1 \frac{\partial G}{\partial \mathbf{A}},
	\end{equation}
	where
	\begin{equation}
	\label{coupled2}
	(\frac{\partial G}{\partial \mathbf{A}})^T = \mathbf{x}^t(\mathbf{A}\mathbf{x}^{t}-\mathbf{b})^T  = \mathbf{x}^t\hat{G}(\mathbf{A},\mathbf{x}).
	\end{equation}
	The  function $\hat{G}$ is defined by considering the bilinear optimization problem  as in Eq.~\ref{eq:bl_sparse}, and we have
    \begin{equation}
    	\label{gacap}
    \hat{G}(A,x) = (\mathbf{A}\mathbf{x}^{t}-\mathbf{b})^T.
    \end{equation}

    Eq. \ref{coupled2}  shows that the gradient for $\mathbf{A}$ tends to vanish,  when $\mathbf{x}$ approaches zero due to the sparsity regularization term $||x||_1$.  Although it has a chance to be  corrected in some  tasks, more likely,  the  update will cause an asynchronous convergence.  Note that for simplicity, the regularization term on $A$ is not considered. Similarly, for $\mathbf{x}$, we have
	\begin{equation}
	\label{bch1}
	\mathbf{x}^{t+1} = \mathbf{x}^{t} + \eta_2 \frac{\partial G}{\partial \mathbf{x}}.
	\end{equation}
     $\eta_1$ and $\eta_2$ are the learning rates.	The conventional gradient descent algorithm for bilinear models iteratively optimizes one variable while keeping the other fixed. This unfortunately ignore the relationship of the two hidden variables in optimization. 
	
	\subsection{Cogradient Descent for Dependable Learning}
	\label{sec:theory}
	We consider the problem from a new perspective such that $\mathbf{A}$ and $\mathbf{x}$ are coupled. 
	Firstly, based on the chain rule~\cite{petersen2008matrix} and its notations, we have
	\begin{equation}
	\begin{aligned}
	\label{coupled1}
	\hat{x}^{t+1}_j &= {x}^{t}_j + \eta_2(\frac{\partial G}{\partial x_j} + Tr((\frac{\partial {G}}{\partial \mathbf{A}})^T \frac{\partial {\mathbf{A}}}{\partial x_j})), \\
	\end{aligned}
	\end{equation}
	where 
	$(\frac{\partial G}{\partial \mathbf{A}})^T = \mathbf{x}^t\hat{G}(\mathbf{A},\mathbf{x})$ as shown in Eq. \ref{coupled2}. $Tr(\cdot)$ represents the trace of the  matrix, which means that each
	element in the matrix $\frac{\partial G}{\partial x_j}$ adds the trace of the corresponding matrix related to $x_j$.
	Considering
	\begin{equation}
		\frac{\partial G}{\partial \mathbf{A}}=\mathbf{A}\mathbf{x}^t\mathbf{x}^{T,t}-b\mathbf{x}^{T,t},
	\end{equation}	
	we have
	\begin{equation}
	\begin{aligned}
	\frac{\partial G(\mathbf{A})}{\partial x_j}&=Tr[(\mathbf{A}\mathbf{x}^t\mathbf{x}^{T,t}-b\mathbf{x}^{T,t})^T \frac{\partial \mathbf{A}}{\partial x_j}]\\
	& =Tr[((\mathbf{A}\mathbf{x}^t-b)\mathbf{x}^{T,t})^T] \frac{\partial \mathbf{A}}{\partial x_j}\\
	& =Tr[\mathbf{x}^t \hat{G} \frac{\partial \mathbf{A}}{\partial x_j}],
	\end{aligned}
	\label{eq:6}
	\end{equation}
	where $\hat{G}=(\mathbf{A}\mathbf{x}^t-b)^T=[\hat{g}_1,...,\hat{g}_M]$.
	Supposing that $\mathbf{A}_i$ and $x_j$ are independent when $i\ne j$, we have
	\begin{equation}
		\frac{\partial \mathbf{A}}{\partial x_j}=
		\begin{bmatrix} 
		0 &...&\frac{\partial \mathbf{A}_{1j}}{\partial x_j}&...& 0 \\
		.&&.&&.\\
		.&&.&&.\\
		.&&.&&.\\ 
		0 &...&\frac{\partial \mathbf{A}_{Mj}}{\partial x_j}&...& 0
		\end{bmatrix},
		\label{eq:7}
	\end{equation} 
	and
	\begin{equation}
		\mathbf{x}\hat{G}=
		\begin{bmatrix} 
		x_1\hat{g}_1 &...&x_1\hat{g}_j&...& x_1\hat{g}_M \\
		.&&.&&.\\
		.&&.&&.\\
		.&&.&&.\\ 
		x_N\hat{g}_1 &...&x_N\hat{g}_j&...& x_N\hat{g}_M
		\end{bmatrix}.
		\label{eq:8}
	\end{equation}
	Combining Eq.~\ref{eq:7} and Eq.~\ref{eq:8}, we have
	\begin{equation}
		\mathbf{x}\hat{G}\frac{\partial \mathbf{A}}{\partial x_j}=
		\begin{bmatrix} 
		0 &...&x_1\sum_i^M\hat{g}_i\frac{\partial \mathbf{A}_{ij}}{\partial x_j}&...& 0 \\
		.&&.&&.\\
		.&&.&&.\\
		.&&.&&.\\ 
		0 &...&x_N\sum_i^M\hat{g}_i\frac{\partial \mathbf{A}_{ij}}{\partial x_j}&...& 0
		\end{bmatrix}.
		\label{eq:9}
	\end{equation}
	The trace of Eq.~\ref{eq:9} is then calculated by:
	\begin{equation}
		Tr[\mathbf{x}^t \hat{G} \frac{\partial \mathbf{A}}{\partial x_j}]=x_j\sum_i^M\hat{g}_i\frac{\partial \mathbf{A}_{ij}}{\partial x_j}.
		\label{eq:10}
	\end{equation}
	Remembering that $\mathbf{x}^{t+1} = \mathbf{x}^{t} + \eta_2 \frac{\partial G}{\partial \mathbf{x}}$, CoGD is established by combining Eq.~\ref{coupled1} and Eq.~\ref{eq:10}:
	
	\begin{equation}
		\begin{aligned}
		\hat{\mathbf{x}}^{t+1}&=\mathbf{x}^{t+1}+\eta_2
		\begin{bmatrix} 
		\sum_i^M\hat{g}_i\frac{\partial \mathbf{A}_{i1}}{\partial x_1}\\
		.\\
		.\\
		.\\ 
		\sum_i^M\hat{g}_i\frac{\partial \mathbf{A}_{iN}}{\partial x_N}
		\end{bmatrix}    
		\odot
		\begin{bmatrix} 
		x_1\\
		.\\
		.\\
		.\\ 
		x_N
		\end{bmatrix}\\    
		&=\mathbf{x}^{t+1}+\eta_2
		\begin{bmatrix} 
		<\hat{G},\frac{\partial \mathbf{A}_{1}}{\partial x_1}>\\
		.\\
		.\\
		.\\ 
		<\hat{G},\frac{\partial \mathbf{A}_{N}}{\partial x_N}>
		\end{bmatrix}    
		\odot
		\begin{bmatrix} 
		x_1\\
		.\\
		.\\
		.\\ 
		x_N
		\end{bmatrix}\\    
		&=\mathbf{x}^{t+1}+\eta_2 {\mathbf{c}} \odot \mathbf{x}^t.	
		\end{aligned}
	\label{final}
	\end{equation}
	
We further define the kernelized version of $\mathbf{c}$, and have     
	\begin{equation}
	\label{bczhangkernel}
		\begin{aligned}
		{\mathbf{c}} =	 \begin{bmatrix} 
	\hat{K}(\hat{G},\frac{\partial \mathbf{A}_{1}}{\partial x_1})\\
		.\\
		.\\
		.\\ 
		\hat{K}(\hat{G},\frac{\partial \mathbf{A}_{N}}{\partial x_N})
		\end{bmatrix},   
			\end{aligned}
		\end{equation}
	where $\hat{K}(.,.)$ is a kernel function\footnote{$\hat{K}(x1,x2) = (x1\cdot x2 )^k $}. 	
	Remembering that Eq. \ref{bch1}, $\mathbf{x}^{t+1} = \mathbf{x}^{t} + \eta_2 \frac{\partial G}{\partial \mathbf{x}}$, Eq.~\ref{coupled1} then becomes 
	\begin{equation}
	\label{coupled4}
	\hat{\mathbf{x}}^{t+1} = \mathbf{x}^{t+1} +  \eta_2 \mathbf{c}^{t} \odot \mathbf{x}^{t},
	\end{equation}
	where $\odot$ represents the Hadamard product.   
    It is then  reformulated as a  projection function as
	\begin{equation}
	\label{coupled5}
	\hat{\mathbf{x}}^{t+1} = P(\mathbf{x}^{t+1},\mathbf{x}^{t})=\mathbf{x}^{t+1} +  \beta \odot  \mathbf{x}^{t},
	\end{equation}
	 which shows the rationality of our method, $i.e.$, it is based on a projection function to solve the asynchronous problem of the bilinear optimization by controlling $\beta$.
	
	We first judge when an asynchronous convergence happens in the optimization based on a form of logical operation as
	\begin{equation}
	\label{bch2}
	(\neg s(\mathbf{x}))\wedge(s(\mathbf{A}))=1,
	\end{equation}
	and
	\begin{equation}
	\label{sign_function}
	s(*)=
	\begin{cases}
	1& if\ R(*)\ \ge \alpha,\\
	0& otherwise,
	\end{cases}
	\end{equation}
	where $\alpha$ represents the threshold which changes for different applications. Eq. \ref{bch2} describes an assumption that an asynchronous convergence happens for $\mathbf{A}$ and $\mathbf{x}$ when their norms become significantly different. 
	Accordingly, the update rule of the proposed CoGD is defined as
	\begin{equation}
	\hat{\mathbf{x}}^{t+1}=
	\begin{cases}
	P(\mathbf{x}^{t+1},\mathbf{x}^{t})& if\ (\neg s(\mathbf{x}))\wedge(s(\mathbf{A}))=1,\\
	\mathbf{x}^{t+1}& otherwise,
	\end{cases}
	\label{bt_x}
	\end{equation}
	which leads to a synchronous convergence and  generalizes the conventional gradient descent method. CoGD is then established.
	
	Note that  $c$ in Eq. \ref{final} is calculated based on $\hat G$, which differs for applications.
	$\frac{\partial \mathbf{A}_j}{\partial {x_j}} \approx  \frac{\Delta \mathbf{A}_j}{\Delta x_j}$, where $\Delta$ denotes the difference of the variable over the epoch related to the convergence speed. 
	$\frac{\partial \mathbf{A}_j}{\partial {x_j}} =\mathbf{1}$, if $\Delta x_j$ or $x_j$  approaches to zero. 
	With above derivation, we define CoGD within the gradient descent framework, providing a solid foundation for the convergence analysis of CoGD. Based on CoGD, the variables are sufficiently trained and decoupled,  which can enhance the causality of the learning system \cite{causality}. 
	

	\section{Applications}
	We apply the proposed algorithm on Convolutional Sparse Coding (CSC), deep learning to validate its general applicability to bilinear problems including image inpainting, image reconstruction,  network pruning and CNN model training.
	
	\subsection{Convolutional Sparse Coding}
	CSC operates on the whole image, decomposing a global dictionary and set of features. The CSC problem theoretically more challenging than the patch-based sparse coding~\cite{mairal2010online} and requires more sophisticated optimization model.	The reconstruction process is usually based on a bilinear optimization model formulated as
	\begin{equation}
	\begin{aligned}
	\mathop {\arg \min }\limits_{\mathbf{A},\mathbf{x}} 
	&\frac{1}{2}\left\|\mathbf{b} - \mathbf{A}\mathbf{x} \right\|^2_F+\lambda\left\| \mathbf{x} \right\|_1\\
	\textit{s.t.} &\left\| \mathbf{A}_k \right\|^2_2 \le 1 \quad \forall k \in  \{1,\dots,K \},
	\end{aligned}
	\label{problem_csc}
	\end{equation}
	where $\mathbf{b}$ denotes input images. 
	
	$\mathbf{x}=[\mathbf{x}_1^T,\dots,\mathbf{x}_K^T]^T$ denotes coefficients under sparsity regularization. $\lambda$ is the sparsity regularization factor. $\mathbf{A}=[\mathbf{A}_1,\dots,\mathbf{A}_K]$ is a concatenation of Toeplitz matrices representing the convolution with the kernel filters $\mathbf{A}_k$, where $K$ is the number of the kernels.

	In Eq.\ \ref{problem_csc}, the optimized objectives or models are influenced by two or more hidden factors that interact to produce the observations. 
	Existing solution tend to decompose the bilinear optimization problem into manageable sub-problems~\cite{parikh2014proximal,heide2015fast}. Without considering the relationship between two hidden factors, however, existing methods suffer from sub-optimal solutions caused by an asynchronous convergence speed of the hidden variables. We attempt to purse an  optimized solution based on the proposed CoGD.
	
	Specifically, we introduce a diagonal or block-diagonal matrix $\mathbf{m}$ to the sparse coding framework defined in~~\cite{gu2015convolutional} and reformulate Eq.~\ref{problem_csc} as

	\begin{equation}
	\small
	\begin{aligned}
	&\mathop {\arg \min }\limits_{\mathbf{A},\mathbf{x}} f_1(\mathbf{A}\mathbf{x})+\sum_{k=1}^K(f_2(\mathbf{x}_k)+f_3(\mathbf{A}_k)),
	\end{aligned}
	\label{problem_csc_sum}
	\end{equation}
	
	\noindent where 
		
	\begin{equation}
	\small
	\begin{aligned}
	f_1(\mathbf{v})&=\frac{1}{2}\left\|\mathbf{b}-\mathbf{m}\mathbf{v}\right\|_F^2,\\
	f_2(\mathbf{v})&=\lambda\left\|\mathbf{v}\right\|_1,\\ 
	f_3(\mathbf{v})&=ind_c(\mathbf{v}).
	\end{aligned}
	\label{eq:proximal_csc}
	\end{equation}
	\noindent In Eq.\ \ref{eq:proximal_csc}, $ind_c(\cdot)$ is an indicator function defined on the convex set of the constraints $C = \{\mathbf{x} \vert  \left\| \mathbf{Sx} \right\|_2^2 \le 1 \}$. 
	Similar to Eq.~\ref{bt_x}, we have
	\begin{equation}
	\small
	\begin{aligned}
	&\ \ \ \ \ \ \hat{\mathbf{x}}_k =
	\begin{cases}
	P({\mathbf{x}}^{t+1}_k,{\mathbf{x}}^{t}_k)& if\ (\neg s(\mathbf{x}_k))\wedge(s(\mathbf{A}_{k}))=1\\
	{\mathbf{x}}^{t+1}_k& otherwise
	\end{cases},
	\end{aligned}
	\label{bt_csc}
	\end{equation}	
	which solve the two coupled variables iteratively, yielding a new CSC solution as defined in Alg.~\ref{alg_csc}.  	$P({\mathbf{x}}^{t+1}_k,{\mathbf{x}}^{t}_k)$ is calculated based on $\hat{G}(\mathbf{A},\mathbf{x})$, which is defined in Eq. \ref{gacap}. 
	
	\input{algorithm/algorithm_CSC.tex}
		
	\begin{figure*}[!t]
		\centering
		\includegraphics[width=1\textwidth]{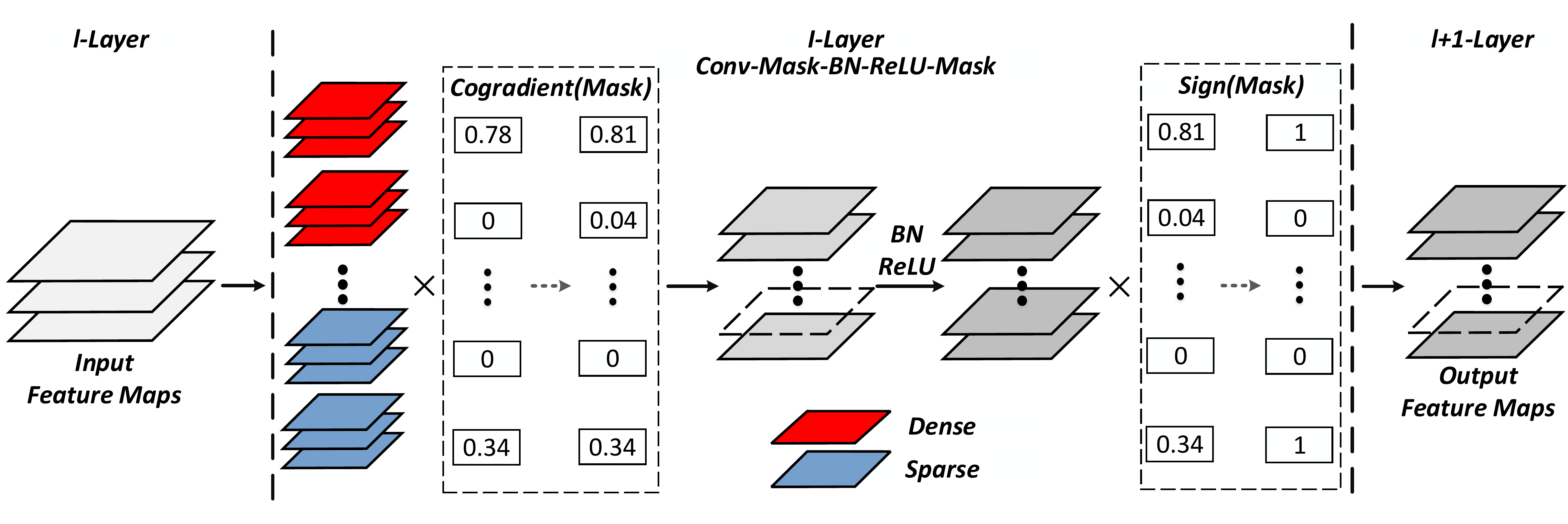}
		\caption{The forward process with the soft mask.}
		\label{mask1}
	\end{figure*}

	\subsection{Network Pruning}
	Network pruning, particularly convolutional channel pruning, has received increased attention for compressing CNNs. Early works in this area tended to directly prune the kernel based on simple criteria like the norm of kernel weights~\cite{li2016pruning} or use a greedy algorithm~\cite{luo2017thinet}. Recent approaches have formulated network pruning as a bilinear optimization problem with soft masks and sparsity regularization~\cite{he2017channel,ye2018rethinking,Huang2017Data,Lin2019Towards}.
	
	Based on the framework of channel pruning~\cite{he2017channel,ye2018rethinking,Huang2017Data,Lin2019Towards}, we apply the proposed CoGD for network pruning. 
	To prune a channel of the network, the soft mask $\mathbf{m}$ is introduced after the convolutional layer to guide the output channel pruning.
	This is defined as a bilinear model  as 
	\begin{equation}
	F_j^{l + 1} =  f(\sum\limits_i {F_i^l \otimes (W_{{{i,j}}}^l}{\mathbf{m}_j}) ),
	\end{equation}
	where \(F_j^{l}\) and \(F_j^{l + 1}\) are the \( i \)-th input and the \( j \)-th output feature maps at the $l$-th and $(l+1)$-th layer. \( W^l_{i,j}\) are convolutional filters corresponding to the soft mask \(\mathbf{m}\). \( \otimes \) and \( f( \cdot )\) respectively refer to convolutional operator and activation.
	
	\input{algorithm/algorithm_pruning}	
	
	In this framework shown in Fig.\ \ref{mask1}, the soft mask $\mathbf{m}$ is learned end-to-end in the back propagation process. To be consistent with other pruning works, we use $W$ and $\mathbf{m}$ instead of $\mathbf{A}$ and $\mathbf{x}$. A general optimization function for network pruning with a soft mask is defined as
	\begin{equation}
	\mathop {\arg \min }\limits_{W,\mathbf{m}}
	\mathcal{L} ( W, {\mathbf{m}}) +\lambda{\left\| \mathbf{m} \right\|_1}+R({W}),
	\label{problem_pruning}
	\end{equation}
	where $\mathcal{L}(W,{\mathbf{m}})$ is the loss function, described in details below. With the sparsity constraint on $\mathbf{m}$, the convolutional filters with zero value in the corresponding soft mask are regarded as useless filters. This means that these filters and their corresponding channels in the feature maps have no significant  contribution to the network predictions and should be pruned. 
	There is, however, a dilemma in the pruning-aware training in that the pruned filters are not evaluated well before they are pruned, which leads to sub-optimal pruning. In particular, the soft mask $\mathbf{m}$ and the corresponding kernels are not sparse in a synchronous manner, which can prune the kernels still of potentials.
	To address this problem, we apply the proposed CoGD to calculate the soft mask $\mathbf{m}$, by reformulating Eq.~\ref{bt_x} as	
	\begin{equation}
	\small
	\begin{aligned}
	\hat{\mathbf{m}}^{l,t+1}_j\!=&
	\begin{cases}
	P({\mathbf{m}_j}^{l,t+1},{\mathbf{m}_j}^{l,t})&if\ (\neg s(\mathbf{m}_j^{l,t}))\wedge s(\sum_i W_{i,j}^l)\!=\!1\\
	{\mathbf{m}}^{l,t+1}_j & otherwise,
	\end{cases}
	\end{aligned}
    \label{bt_pruning}
	\end{equation}
	where $W_{i,j}$ represents the 2D kernel of the  $i$-th input channel of the $j$-th filter. {$\beta$, $\alpha_{W}$  and $\alpha_{\mathbf{m}}$ are detailed in experiments.}
The form of $\hat{G}$ is specific for different applications. For CNN pruning,  based on Eq.~\ref{coupled2}, we simplify the calculation of $\hat{G}$ as
	\begin{equation}
	\hat{G} = \frac{\partial \mathcal{L}}{\partial W_{i,j}}/\mathbf{m}_j.
	\label{eq_hatg}
	\end{equation}	
Note that the the autograd package in deep learning frameworks such as Pytorch \cite{paszke2019pytorch} can automatically calculate $\frac{\partial \mathcal{L}}{\partial W_{i,j}}$.  	We then Substitute Eq. \ref{eq_hatg} into Eq.  \ref{bczhangkernel} to train our network, and prune CNNs based on the new mask $\hat{\mathbf{m}}$ in  Eq.~\ref{bt_pruning}. 
	
To examine how our CoGD works for network pruning, we use GAL \cite{Lin2019Towards} as an example to describe our CoGD for CNN pruning.
	A pruned network obtained through GAL with \(\ell_{1}\)-regularization on the soft mask is used to approximate the pre-trained network by aligning their outputs. The discriminator \(D\) with weights \(W_D\) is introduced to discriminate between the output of the pre-trained network and the pruned network. The pruned network generator  \(G\) with weights \(W_G\) and soft mask $\mathbf{m}$ is learned together with \(D\) by using the knowledge from supervised features of the baseline.    
	Accordingly, the soft mask \(\mathbf{m}\), the new mask \(\hat{\mathbf{m}}\), the pruned network weights \(W_G\), and the discriminator weights \(W_D\) are simultaneously learned by solving the optimization problem as follows: 
	\begin{equation}
	\small
	\begin{aligned}
    	\arg \mathop {\min }\limits_{{W_G},\mathbf{m}} \mathop {\max }\limits_{{W_D},\hat{\mathbf{m}}} &{\mathcal{L}_{Adv}}({W_G},\hat{\mathbf{m}},{W_D})+ {\mathcal{L}_{data}}({W_G},\hat{\mathbf{m}})\\
	+ &{\mathcal{L}_{reg}}({W_G},\mathbf{m},{W_D}).
	\end{aligned}
	\label{optimization}
	\end{equation}    
	where $\mathcal{L} ( W, {\mathbf{m}}) = {\mathcal{L}_{Adv}}({W_G},\hat{\mathbf{m}},{W_D})+ {\mathcal{L}_{data}}({W_G},\hat{\mathbf{m}})$ and 	 ${\mathcal{L}_{reg}}({W_G},\mathbf{m},{W_D})$ are related to $\lambda{\left\| \mathbf{m} \right\|_1}+R({W})$ in Eq. \ref{problem_pruning}.  \({\mathcal{L}_{Adv}}({W_G},\hat{\mathbf{m}},{W_D})\) is the adversarial loss to train the two-player game between the pre-trained network and the pruned network that compete with each other. Details of the algorithm are described in Alg.~\ref{alg_gal}.

	\begin{figure}[!t]
		\centering
		\includegraphics[width=0.7\linewidth]{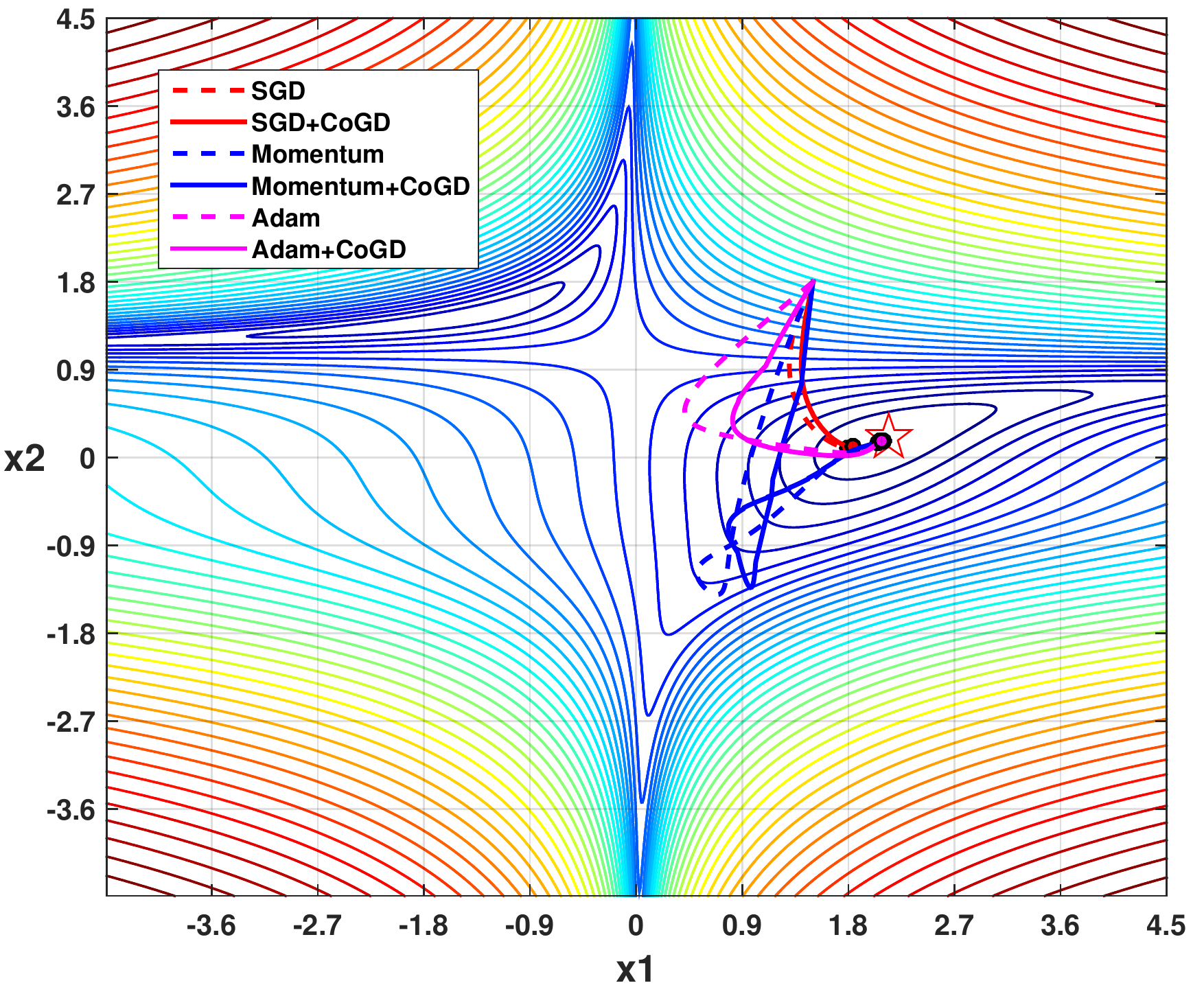}%
		\caption{ Comparison of classical gradient algorithm with CoGD.The background is the contour map of Beale functions. The algorithms with CoGD have short optimization paths compared with their counterparts, which shows that CoGD facilitates efficient and sufficient training.}
		\label{optimizer}
	\end{figure}
	
	The advantages of CoGD in network pruning are three-fold. First, CoGD that optimizes the bilinear pruning model leads to a synchronous gradient convergence. Second, the process is controllable by a threshold, which makes the pruning rate easy to adjust. Third, the CoGD method for network pruning is scalable, $i.e.$, it can be built upon other state-of-the-art networks for better performance. 
	
	\subsection{CNN Training} 
	The last but not the least, CoGD can be fused with the Batch Normalization (BN) layer and improve the performance of CNN models.	As is known, the BN layer can re-distribute the features, resulting that the feature and kernel learning converge in an asynchronous manner.
	CoGD is then introduced to synchronize their learning speeds to sufficiently train CNN models. 
	In specific, we backtrack sparse convolutional kernels through evaluating the sparsity of the BN layer, leading to an efficient training process. 
    To ease presentation, we first copy Eq. \ref{eq_convbc} as
     \begin{equation}
	    F_j^{l + 1} =  f(BN(\sum\limits_i {F_i^l \otimes W_{{{i,j}}}^l})),
	    \label{eq_convbc1},
	    \end{equation}
	    then redefine the BN model as
    \begin{equation}
	\begin{aligned}
	BN(x) = \gamma \frac{x - \mu_B}{\sqrt{\sigma_B + \epsilon}} + \beta, \\
	\mu_B=\frac{1}{m}\sum_{i=1}^{m}{x_i},\\
	\sigma_B=\frac{1}{m}\sum_{i=1}^{m}(x_i-\mu_B)^2,
	\end{aligned}
	\label{eq:BN}
	\end{equation}
    where $m$ is the mini-batch size, $\mu_B$ and $\sigma_B$ are mean and variance obtained by feature calculation in the BN layer. $\gamma$ and $\beta$ are the learnable parameters, and $\epsilon$ is a small number to avoid dividing by zero. 
    
    According to Eq. \ref{eq_convbc1} and Eq. \ref{eq:BN}, we can easily know that $\gamma$ and $W$ are bilinear. We use the sparsity of $\gamma$ instead of the whole convolutional features for kernels backtracking, which  simplifies the operation and improves the backtracking efficiency. Similar to network pruning, we also use $\gamma$ and $W$ instead of $\mathbf{A}$ and $\mathbf{x}$ in this part. A general optimization  for CNN training with the BN layer  as

	\begin{equation}
	\mathop {\arg \min }\limits_{W, \gamma}
	\mathcal{L} (W, \gamma) +\lambda{\left\| W \right\|_1}, 
	\label{problem_cnnlearing}
	\end{equation}	
	where $\mathcal{L}(W, \gamma)$ is the loss function defined on Eq. \ref{eq_convbc1} and Eq. \ref{eq:BN}. CoGD is then applied to train CNNs, by reformulating Eq.~\ref{bt_x} as
	
	\begin{equation}
	\small
	\begin{aligned}
	\hat{W}^{l,t+1}_{i,j}\!=&
	\begin{cases}
	P({W_{i,j}}^{l,t+1},{W_{i,j}}^{l,t})&if\ (\neg s(\gamma_{j}^l))\wedge s(\sum_i W_{i,j}^l)\!=\!1\\
	{W_{i,j}}^{l,t} & otherwise,
	\end{cases}
	\end{aligned}
    \label{bt_cnnlearning}
	\end{equation}	
	where $\gamma_{j}^l$ is the $j$-th learnable parameter in the $l$-th BN layer. $W_{i,j}$ represents the 2D kernel of the  $i$-th input channel of the $j$-th filter.  Similar to network pruning, we define
	
	\begin{equation}
	\hat{G} = \frac{\partial \mathcal{L}}{\partial W_{i,j}}/\mathbf{\gamma}_j,
	\label{gcapcnn}
	\end{equation}
    where $\frac{\partial \mathcal{L}}{\partial W_{i,j}}$ is obtained based on  the autograd package in deep learning frameworks such as Pytorch \cite{paszke2019pytorch}.   Similar to network pruning, we substitute Eq. \ref{gcapcnn} into Eq. \ref{bczhangkernel}, to use CoGD  for CNN  training, yielding Alg. \ref{alg_cnn}.

\input{algorithm/cnntraining.tex}

	

\section{Experiments}
    In this section, CoGD is first analyzed and compared with classical optimization methods on a baseline problem. It is then validated on the problems of CSC, network pruning, and CNN model training.

	\subsection{Baseline Problem}
	A baseline problem is first used as an example to illustrate the superiority of our algorithm. 
	The problem is the optimization of Beale function \footnote{$beale(x_1,x_2)=(1.5-x_1+x_1x_2)^2+(2.25-x_1+x_1x_2^2)^2+(2.62-x_1+x_1x_2^3)^2$.} under  constraint of $F(x_1,x_2)=beale(x_1,x_2)+\left\|x_1\right\|+x_2^2$.
	The Beale function has the same form as Eq.~\ref{eq:bl_sparse} and can be regraded as a bilinear optimization problem with respect to variables $x_1 x_2$.
	During optimization, the learning rate $\eta_2$ is set as $0.001$, $0.005$, $0.1$ for `SGD', `Momentum' and `Adam' respectively. The thresholds $\alpha_{x_1}$ and $\alpha_{x_2}$  for CoGD are set to $1$ and $0.5$. $\beta = 0.001 \eta_2 \mathbf{c}^{t}$ with $\frac{\partial x_2}{\partial {x_1}} \approx  \frac{\Delta x_2}{\Delta x_1}$, where $\Delta$ denotes the difference of variable over the epoch. $\frac{\Delta x_2}{\Delta x_1} =\mathbf{1}$, when $\Delta x_2$ or $x_2$  approaches zero. The total  number of iterations is $200$.

	In Fig.~\ref{optimizer}, we compare the optimization paths of CoGD with those of three widely used optimization methods - `SGD', `Momentum' and `Adam'. It can be seen that algorithms equipped with CoGD have shorter optimization paths than their counterparts. Particularly, the ADAM-CoGD algorithm has a much shorter path than ADAM, demonstrating the fast convergence of the proposed CoGD algorithm. The similar convergence with shorter paths means that CoGD facilitates efficient and sufficient training. 
	 
	

	\begin{figure}[!t]
		\centering
		\includegraphics[width=0.7\linewidth]{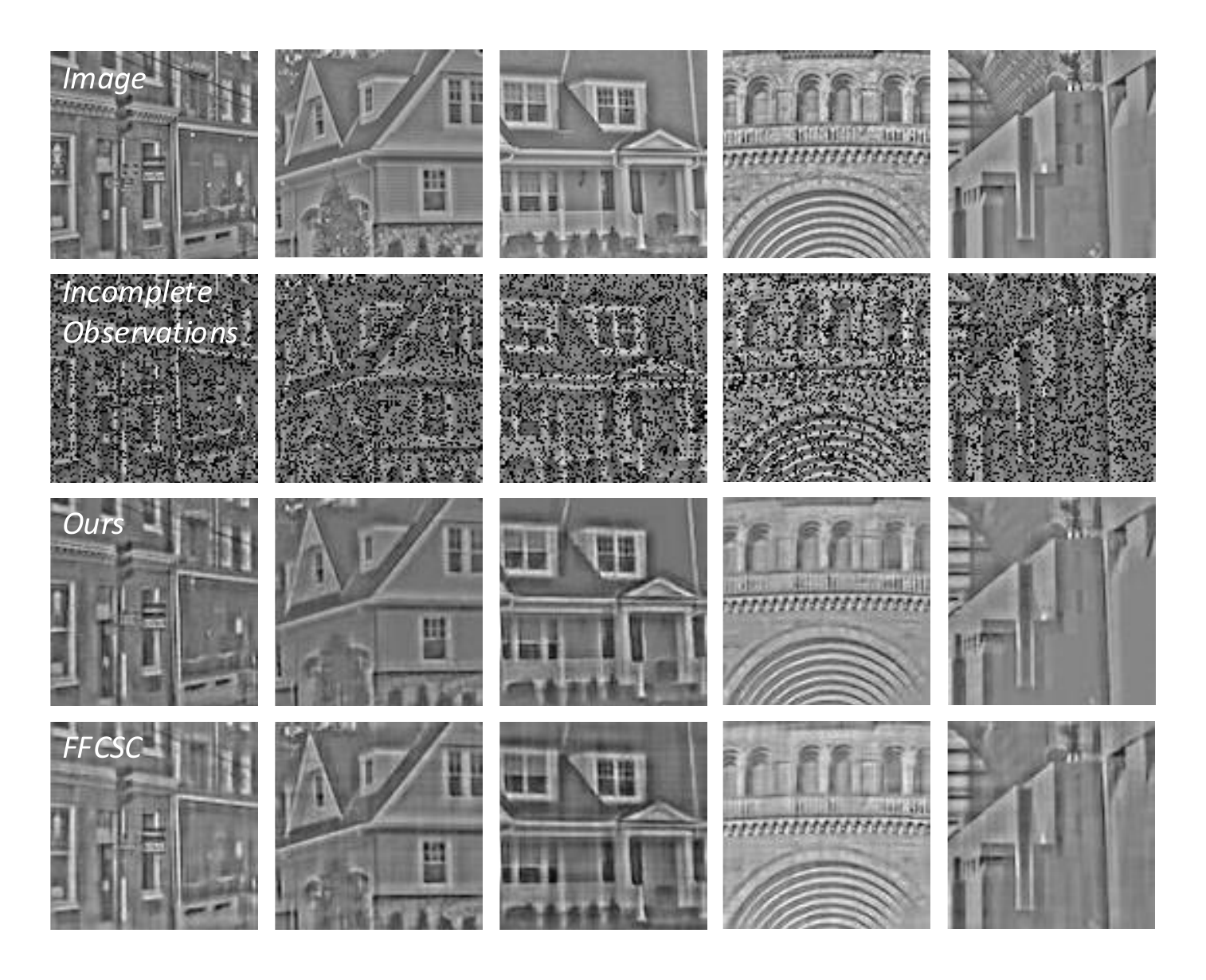}
		\caption{Inpainting for the normalized city dataset. From top to bottom: the original images, incomplete observations, reconstructions with FFCSC~\cite{heide2015fast},  and reconstructions with our proposed algorithm. }
		\label{Inpainting}
	\end{figure}

	\subsection{Convolutional Sparse Coding}
	
	\textbf{Experimental Setting.} The CoGD for convolutional sparse coding (CSC) is evaluated on two  public datasets: the fruit dataset~\cite{zeiler2010deconvolutional} and the city dataset~\cite{zeiler2010deconvolutional,heide2015fast}, each of which consists of ten images with 100 $\times$ 100 resolution. To evaluate the quality of the reconstructed images, we use two standard metrics, the peak signal-to-noise ratio (PSNR, unit:dB) and the structural similarity (SSIM). The higher the PSNR and the SSIM values are, the better the visual quality of the reconstructed image is. The evaluation metrics are defined as:
	\begin{equation}	
	PSNR = 10 \times \log_{10}  (\frac{MAX^2}{MSR}), 
	\label{eq_psnr}
	\end{equation}
    where $MSE$ is the mean square error of clean image and noisy image. $MAX$ is the maximum pixel value of the image.

	\begin{equation}	
	SSIM(x,y) =  \frac{(2\mu_x\mu_y+C_1)(2\sigma_{xy}+C_2)}{(mu_x^2+\mu_y^2+C_1)(\sigma_x^2+\sigma_y^2+C_2)},
	\label{eq_ssim}
	\end{equation}
    where $\mu$ is the mean of samples. $\sigma$ is the variance of samples. $\sigma_{xy}$ is the covariance of the samples. $C$ is a constant, $C_1=(0.01\times MAX)^2$, $C_2=(0.03\times MAX)^2$
	
	\textbf{Implementation Details:} The reconstruction model is implemented based on the conventional CSC method~\cite{gu2015convolutional}, while we introduce the CoGD with the kernelized projection function to achieve a better convergence and higher reconstruction accuracy. One hundred of filters with size 11$\times$11 are set as model parameters. $\alpha_{\mathbf{x}}$ is set to the mean of $\left\|\mathbf{x}_k\right\|_1$. $\alpha_{\mathbf{A}}$ is calculated as the median of the sorted results of $\beta_k$. 
	As shown in Eq. \ref{bczhangkernel}, linear and polynomial kernel functions are used in the experiment, which can both improve the performance of our method. For a fair comparison, we use the same hyperparameters ($\eta_2$) in both our method and~\cite{gu2015convolutional}. We also test $\beta = 0.1 \eta_2 \mathbf{c}^{t}$, which achieves a similar performance as the linear kernel. 
	
	\begin{table*}[!t]
		\centering
		\small
		\caption{Inpainting results for convolutional filters learned with the proposed method and with FFCSC~\cite{heide2015fast}. All reconstructions are performed with 75\% data subsampling. The proposed CoGD achieves better PSNR and SSIM in all cases.}
		\scalebox{0.7}{
			\begin{tabular}{c c c c c c c c c c c c c c}
				\hline\hline
				Dataset&Fruit& 1 & 2 & 3 & 4 & 5 & 6 & 7 & 8 & 9 & 10 & Average \\ \hline
				\multirow{3}{*}{PSNR (dB)}&~\cite{heide2015fast} & 25.37 & 24.31 & 25.08 & 24.27 & 23.09 & 25.51 & 22.74& 24.10 & 19.47 & 22.58 & 23.65 \\ 
				&CoGD(kernelized, $k=1$) & 26.37 & 24.45 & 25.19 & 25.43 & 24.91 & \textbf{27.90} & 24.26 & 25.40 & \textbf{24.70} & \textbf{24.46} & 25.31 \\ 
				&CoGD(kernelized, $k=2$) & 27.93 & \textbf{26.73} & 27.19 & 25.25 & \textbf{23.54} & 25.02 & \textbf{26.29} & 24.12 & 24.48 & 24.04 & 25.47 \\
				&CoGD(kernelized, $k=3$) & \textbf{28.85} & 26.41 & \textbf{27.35} & \textbf{25.68} & 24.44 & 26.91 & 25.56 & \textbf{25.46} & 24.51 & 22.42 & \textbf{25.76} \\ 
				\hline	
				\multirow{3}{*}{SSIM}&~\cite{heide2015fast} & 0.9118 & 0.9036 & 0.9043 & 0.8975 & 0.8883 & 0.9242 & 0.8921 & 0.8899 & 0.8909 & 0.8974 & 0.9000 \\ 
				&CoGD(kernelized, $k=1$) & 0.9452 & 0.9217 & \textbf{0.9348} & 0.9114 & \textbf{0.9036} & \textbf{0.9483} & 0.9109 & 0.9041 & \textbf{0.9215} & \textbf{0.9097} & \textbf{0.9211} \\ 
				&CoGD(kernelized, $k=2$) & 0.9483 & \textbf{0.9301} & 0.9294 & 0.9061 & 0.8939 & 0.9454 & \textbf{0.9245} & 0.8990 & 0.9208 & 0.9054 & 0.9203 \\
				&CoGD(kernelized, $k=3$)& \textbf{0.9490} & 0.9222 & 0.9342 & \textbf{0.9181} & 0.8810 & 0.9464 & 0.9137 & \textbf{0.9072} & 0.9175 & 0.8782 & 0.9168 \\  				
				\hline\hline		
				
				Dataset&City& 1 & 2 & 3 & 4 & 5 & 6 & 7 & 8 & 9 & 10 & Average \\ \hline
				\multirow{3}{*}{PSNR (dB)}&~\cite{heide2015fast} & 26.55 & 24.48 & 25.45 & 21.82 & 24.29 & 25.65 & 19.11 & 25.52 & 22.67 & 27.51 & 24.31\\ 
				&CoGD(kernelized, $k=1$)  & 26.58& 25.75& 26.36 & 25.06 & 26.57& 24.55& 21.45& \textbf{26.13}& 24.71& \textbf{28.66}& 25.58  \\ 
				&CoGD(kernelized, $k=2$) & \textbf{27.93} & \textbf{26.73} & \textbf{27.19} & 25.83 & 24.41 & 25.31 & \textbf{26.29} & 24.70 & 24.48 & 24.62 & \textbf{25.76} \\
				&CoGD(kernelized, $k=3$) & 25.91 & 25.95 & 25.21 & \textbf{26.26} & \textbf{26.63} & \textbf{27.68} & 21.54 & 25.86 & \textbf{24.74} & 27.69 & 25.75 \\ 			
				\hline
				\multirow{3}{*}{SSIM}&~\cite{heide2015fast} & 0.9284 & 0.9204 & 0.9368 & 0.9056 & 0.9193 & 0.9202 & 0.9140 & 0.9258 & 0.9027 & 0.9261 & 0.9199 \\ 
				&CoGD(kernelized, $k=1$)  & 0.9397& 0.9269& \textbf{0.9433}& \textbf{0.9289}& 0.9350& 0.9217& \textbf{0.9411}& 0.9298& 0.9111& 0.9365& 0.9314  \\ 
				&CoGD(kernelized, $k=2$) & \textbf{0.9498} & \textbf{0.9316} & 0.9409 & 0.9176 & 0.9189 & \textbf{0.9454} & 0.9360 & 0.9305 & \textbf{0.9323} & 0.9284 & 0.9318 \\
				&CoGD(kernelized, $k=3$) & 0.9372 & 0.9291 & 0.9429 & 0.9254 & \textbf{0.9361} & 0.9333 & 0.9373 & \textbf{0.9331} & 0.9178 & \textbf{0.9372} & \textbf{0.9329} \\  	 				
				\hline\hline
		\end{tabular}}
		\label{tab:inpainting}
	\end{table*}	
	
	\begin{table*}[!t]
		\centering
		\small
		\caption{Reconstruction results for filters learned with the proposed method and with FFCSC~\cite{heide2015fast}. With the exception of 6 images, the proposed method achieves better PSNR and SSIM.}
		\scalebox{0.7}{
			\begin{tabular}{c c c c c c c c c c c c c c}
				\hline\hline
				Dataset&Fruit& 1 & 2 & 3 & 4 & 5 & 6 & 7 & 8 & 9 & 10 & Average \\ \hline
				\multirow{3}{*}{PSNR (dB)}&~\cite{heide2015fast} & 30.90 & \textbf{29.52} & 26.90 & 28.09 & 22.25 & 27.93 & 27.10& 27.05 & 23.65 & 23.65 & 26.70 \\ 
				&CoGD(kernelized, $k=1$) & \textbf{31.46} & 29.12 & 27.26 & \textbf{28.80} & 25.21 & 27.35 & 26.25 & \textbf{27.48} & \textbf{25.30} & \textbf{27.84} & 27.60 \\ 
				&CoGD(kernelized, $k=2$) &30.54 & 28.77 & \textbf{30.33} & 28.64 & \textbf{25.72} &\textbf{30.31} & \textbf{28.07} & 27.46 & 25.22 & 26.14 & \textbf{28.12} \\
				\hline	
				\multirow{3}{*}{SSIM}& ~\cite{heide2015fast} & 0.9706 & 0.9651 & 0.9625 & 0.9629 & 0.9433 & 0.9712 & 0.9581 & 0.9524 & 0.9608 & 0.9546 & 0.9602 \\ 
				&CoGD(kernelized, $k=1$) & \textbf{0.9731} & 0.9648 & 0.9640 & 0.9607 & \textbf{0.9566} & 0.9717 & 0.9587 & 0.9562 & 0.9642 & \textbf{0.9651} & 0.9635 \\ 
				&CoGD(kernelized, $k=2$) & 0.9705 & \textbf{0.9675} & \textbf{0.9660} & \textbf{0.9640} & 0.9477 & \textbf{0.9728} & \textbf{0.9592} & \textbf{0.9572} & \textbf{0.9648} & 0.9642 & \textbf{0.9679} \\
				\hline\hline		
				
				Dataset&City& 1 & 2 & 3 & 4 & 5 & 6 & 7 & 8 & 9 & 10 & Average \\ \hline
				\multirow{3}{*}{PSNR (dB)}&~\cite{heide2015fast} & 30.11 & 27.86 & \textbf{28.91} & 26.70 & 27.85 & 28.62 & 18.63 & 28.14 & 27.20 & 25.81 & 26.98\\ 
				&CoGD(kernelized, $k=1$) & 30.29 & \textbf{28.77}& 28.51& 26.29& 28.50& \textbf{30.36}& 21.22& \textbf{29.07}& \textbf{27.45}& \textbf{30.54}& \textbf{28.10}  \\ 
				&CoGD(kernelized, $k=2$) & \textbf{30.61} & 28.57 & 27.37 & \textbf{27.66} & \textbf{28.57} &29.87 & \textbf{21.48} & 27.08 & 26.82 & 29.86 &  27.79 \\ 
				\hline
				\multirow{3}{*}{SSIM}&~\cite{heide2015fast} & 0.9704 & \textbf{0.9660} & \textbf{0.9703} & 0.9624 & 0.9619 & 0.9613 & 0.9459 & 0.9647 & 0.9531 & 0.9616 & 0.9618 \\ 
				&CoGD(kernelized, $k=1$) & \textbf{0.9717}& \textbf{0.9660}& 0.9702& \textbf{0.9628}& \textbf{0.9627}& \textbf{0.9624}& \textbf{0.9593}& \textbf{0.9663}& \textbf{0.9571}&\textbf{ 0.9632}& \textbf{0.9642}  \\ 
				&CoGD(kernelized, $k=2$) & 0.9697 & 0.9646 & 0.9681 & 0.962 & 0.9613  & 0.9594 & 0.9541 & 0.9607 & 0.9538 & 0.9620 & 0.9631 \\
				\hline\hline
		\end{tabular}}
		\label{tab:reconstruction}
	\end{table*}
	
	\begin{figure}[!t]
		\centering
		
		\subfloat{
			\label{filters1}
			\includegraphics[width=0.7\linewidth]{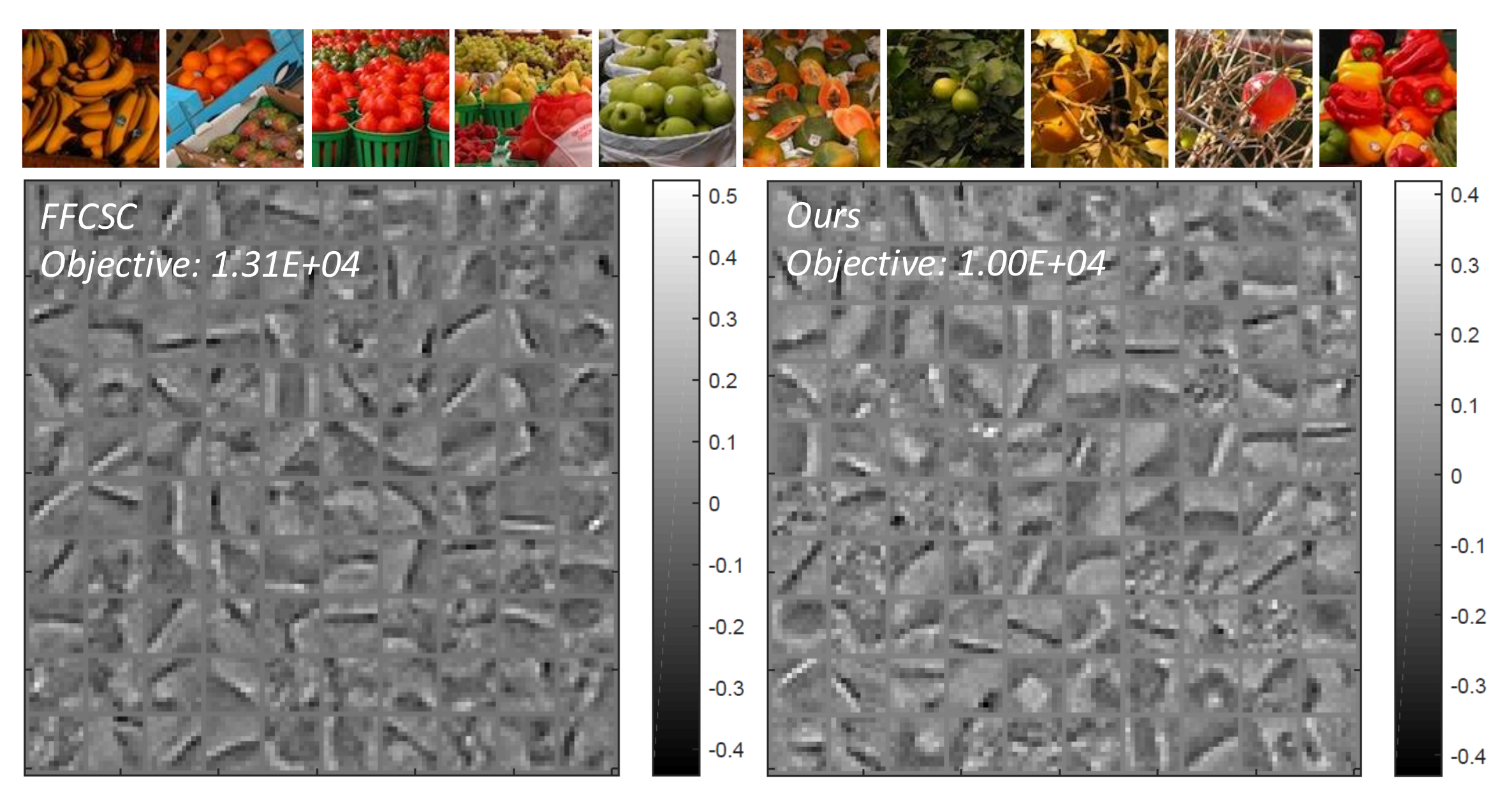}}
		
		\subfloat{
			\label{filters2}
			\includegraphics[width=0.7\linewidth]{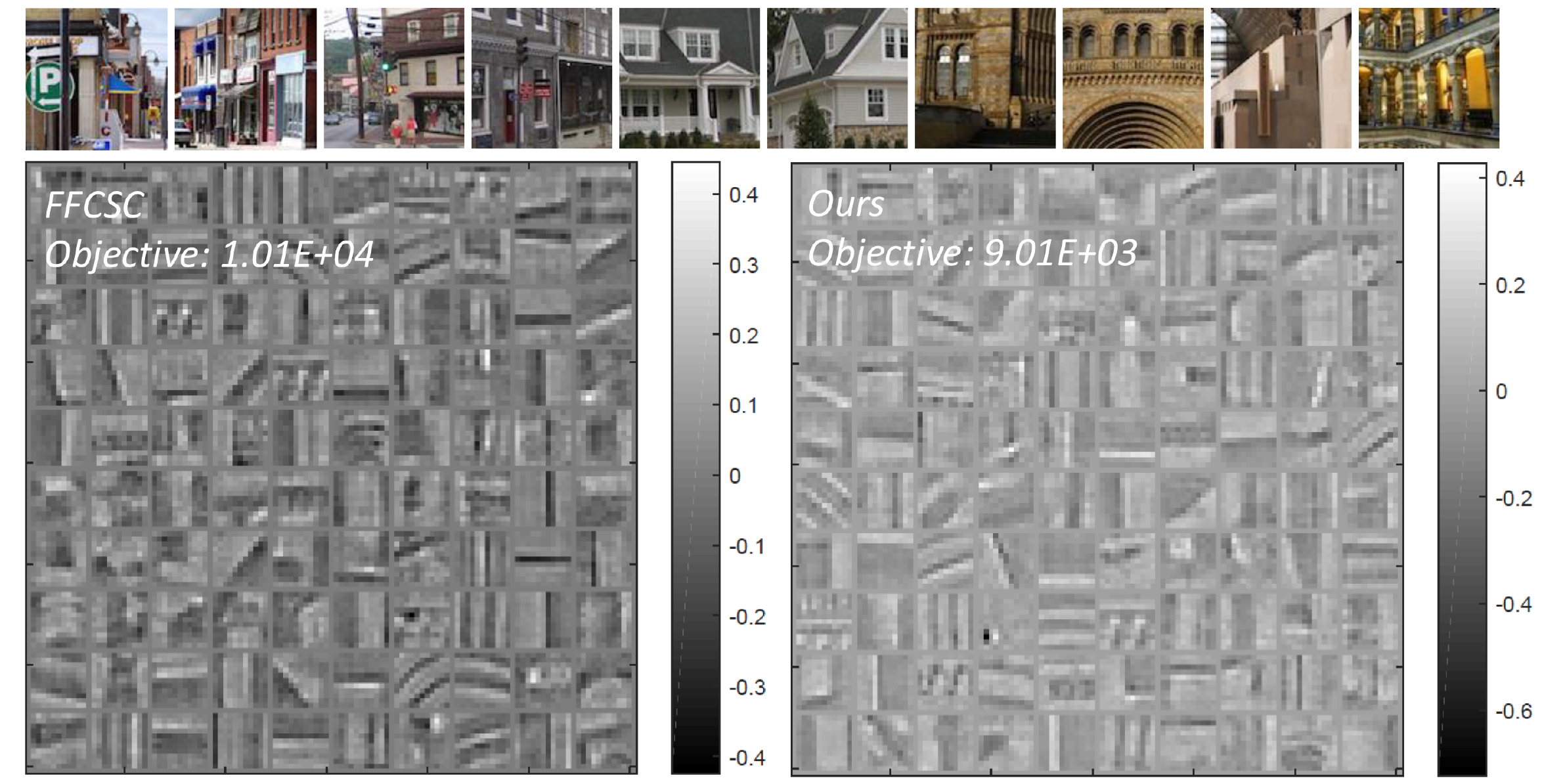}}
		\caption{Filters learned on fruit and city datasets. Thumbnails of the datasets along with filters learned  with FFCSC~\cite{heide2015fast} (left) and with CoGD (right) are shown. The proposed reconstruction method reports lower objectives. (Best viewed in color with zoom)}
		\label{fig:filters}
	\end{figure}	
	
	\textbf{Results:} The CSC with the proposed CoGD algorithm is evaluated with two tasks including image reconstruction and image inpainting.
	
	For image inpainting, we randomly sample the data with a 75\% subsampling rate, to obtain the incomplete data. Like~\cite{heide2015fast}, we test our method on contrast-normalized images. We first learn filters from all the incomplete data under the guidance of the soft mask $\mathbf{m}$, and then reconstruct the incomplete data by fixing the learned filters. We show inpainting results of the normalized data in Fig.~\ref{Inpainting}. Moreover, to compare with FFCSC, inpainting results on the fruit and city datasets are shown in Tab.~\ref{tab:inpainting}. It can be seen that our method achieves a better PSNR and SSIM in all cases while the average PSNR and SSIM improvements are impressive 1.78 db and 0.017. 
	
	For image reconstruction, we reconstruct the images on the fruit and city datasets. One hundred of  11$\times$11 filters are trained and compared with those of FFCSC~\cite{heide2015fast}. Fig.~\ref{fig:filters} shows the resulting filters after convergence within the same 20 iterations. It can be seen that the proposed reconstruction method, driven with CoGD, converges with a lower loss. When comparing the PSNR and the SSIM of our method with FFCSC in Tab.~\ref{tab:reconstruction}, we can see that in most cases our method achieves higher PSNR and SSIM. The average PSNR and SSIM improvements are respectively $1.27$ db and $0.005$.

    Considering that PSNR is calculated with a $\log$ function, the performance improvement shown in Tab.~\ref{tab:inpainting} and Tab.~\ref{tab:reconstruction} are  significant. 
    Such improvements show that the kernelized projection function improves the performance of the algorithm and reveal the nonlinear interaction of the variables. 
	
	\subsection{Network Pruning}
	We have evaluated the proposed CoGD algorithm on network pruning using the CIFAR-10 and ILSVRC12 ImageNet datasets for the image classification tasks. The commonly used ResNets and  MobileNetV2 are used as the backbone networks to get the pruned network models.
	\begin{table}[!t]
		\small
		\caption{Pruning results of ResNet-18/110 and MobilenetV2 on CIFAR-10. M = million (\({10^6}\)).}
		\centering
		\scalebox{0.83}{
			\begin{tabular}{c c c c}
				\hline\hline
				Model     &FLOPs (M) & Reduction & Accuracy/+FT (\%)  \\ \hline 
				ResNet-18\cite{He2016DeepNew}  &555.42 & - &95.31    \\ 
				CoGD-0.5 &274.74 & 0.51$\times$ &95.11/95.30\\ 
				CoGD-0.8 &423.87 & 0.24$\times$ &95.19/\textbf{95.41}\\ 
				\hline							
				ResNet-56\cite{He2016DeepNew}  & 125.49 & -        & 93.26     \\ 

				GAL-0.6\cite{Lin2019Towards}  &78.30    & 0.38$\times$  &92.98/93.38      \\ 
				GAL-0.8\cite{Lin2019Towards}  &49.99    &0.60$\times$     &90.36/91.58      \\ 
				CoGD-0.5 &48.90 & 0.61$\times$  &92.38/92.95        \\ 
				CoGD-0.8 &82.85 & 0.34$\times$  &93.16/\textbf{93.59}   \\ \hline				
				ResNet-110\cite{He2016DeepNew} &252.89 & - &93.68      \\
				GAL-0.1\cite{Lin2019Towards} &205.70 & 0.20$\times$ &92.65/93.59  \\ 
				GAL-0.5\cite{Lin2019Towards} &130.20 & 0.49$\times$ &92.65/92.74  \\ 
				CoGD-0.5 &  95.03  & 0.62$\times$ & 93.31/93.45                  \\ 
				CoGD-0.8 &  135.76 & 0.46$\times$ & 93.42/93.66         \\ 
				\hline	
				MobileNet-V2\cite{Sandler2018MobileNetV2} &91.15 & - & 94.43 \\ 
				CoGD-0.5 &50.10 & 0.45$\times$ &94.25/- \\ 
				\hline\hline					
		\end{tabular}}
		\label{cifar10_res}
	\end{table}
	
	\subsubsection{Experimental Setting}
	
	\textbf{Datasets:} CIFAR-10 is a natural image classification dataset containing a training set of $50,000$ and a testing set of $10,000$ \(32 \times 32\)  color images distributed over ten classes, including airplanes, automobiles, birds, cats, deer, dogs, frogs, horses, ships, and trucks. The ImageNet classification dataset is more challenging due to the significant increase of image categories, image samples, and sample diversity. For the $1,000$ categories of images, there are $1.2$ million images for training and $50$k images for validation. The large data divergence set an ground challenge for the optimization algorithms when pruning network models.	
	
	\textbf{Implementation:}
	We use PyTorch to implement our method with $3$ NVIDIA TITAN V and $2$ Tesla V100 GPUs. The weight decay and the momentum are set to $0.0002$  and  $0.9$ respectively. The hyper-parameter \( \lambda_\mathbf{m}\) is selected through cross-validation in the range $[0.01, 0.1]$ for ResNet and MobileNetv2. The drop rate is set to $0.1$. The other training parameters are described on a per experiment basis.
	
	To better demonstrate our method, we denote CoGD-a as  an approximated pruning rate of $(1-a)$\% for corresponding channels.  $a$  is associated with the threshold $\alpha_W$, which is given  by its sorted result. For example, if $a = 0.5$,  $\alpha_W$ is the median of the sorted result. $\alpha_{\mathbf{m}}$ is set to be $0.5$ for easy implementation. 
	Similarly, $\beta = 0.001 \eta_2 \mathbf{c}^{t}$ with $\frac{\partial \mathbf{W}}{\partial {m_j}} \approx  \frac{\Delta \mathbf{W}}{\Delta m_j}$.
	Note that our training cost is similar to that of \cite{Lin2019Towards}, since we use our method once per epoch without additional cost.

	\subsubsection{CIFAR-10}
	We evaluated the proposed network pruning method on CIFAR-10 for two popular networks, ResNets and MobileNetV2. The stage kernels are set to 64-128-256-512 for ResNet-18  and 16-32-64  for ResNet-110. For all networks, we add a soft mask only after the first convolutional layer within each block to simultaneously prune the output channel of the current convolutional layer and input channel of next convolutional layer.	The mini-batch size is set to be $128$ for $100$ epochs, and the initial learning rate is set to $0.01$, scaled by $0.1$ over $30$ epochs.
	
	\textbf{Fine-tuning:} 
	In the network fine-tuning after pruning, we only reserve the student model. According to the ‘zero’s in each soft mask, we remove the corresponding output channels of the current convolutional layer and corresponding input channels of the  next convolutional layer. We then obtain a pruned network with fewer parameters and that requires fewer FLOPs. We use the same batch size of $256$ for $60$ epochs as in training. The initial learning rate is changed to $0.1$ and scaled by $0.1$ over $15$ epochs. Note that a similar fine-tuning strategy was used in GAL~\cite{Lin2019Towards}.
	
	\textbf{Results:} Two kinds of networks are tested on the CIFAR-10 database - ResNets and MobileNet-V2. In this section, we only test the linear kernel, which  achieves a similar performance as the full-precision model.
	
	Results for \textbf{ResNets} are shown in Tab.~\ref{cifar10_res}. Compared to the pre-trained network for ResNet-18 with $95.31$\% accuracy, CoGD-$0.5$ achieves a \({\rm{0.51}} \times \) FLOPs reduction with neglibilbe accuracy drop \(0.01\% \). 
	Among other structured pruning methods for ResNet-110, 
	CoGD-$0.5$ has a larger FLOPs reduction than GAL-$0.1$ ($95.03M$ v.s. $205.70M$), but with similar accuracy ($93.45$\% v.s. $93.59$\%). These results demonstrate that our method can prune the network efficiently and generate a more compressed model with higher performance.
	
	\begin{table}[!t]
		\caption{Pruning results of ResNet-50 on ImageNet. B means billion (\({10^9}\)).}
		\centering
		\scalebox{0.83}{
			\begin{tabular}{c c c c}
				\hline\hline
				Model      & FLOPs (B) & Reduction & Accuracy/+FT (\%)  \\ \hline
				ResNet-50\cite{He2016DeepNew}  & 4.09 & - & 76.24   \\ 
				ThiNet-50\cite{luo2017thinet}  & 1.71 & 0.58$\times$ & 71.01               \\ 
				ThiNet-30\cite{luo2017thinet}  & 1.10 & 0.73$\times$ & 68.42               \\ 				CP\cite{he2017channel}         & 2.73 & 0.33$\times$ & 72.30               \\ 
				GDP-0.5\cite{lin2018accelerating}  & 1.57 & 0.62$\times$ & 69.58               \\ 
				GDP-0.6\cite{lin2018accelerating}  & 1.88 & 0.54$\times$ & 71.19               \\ 
				SSS-26\cite{huang2018data}     & 2.33 & 0.43$\times$ & 71.82                \\ 
				SSS-32\cite{huang2018data}     & 2.82 & 0.31$\times$ & 74.18                \\ 
				RBP\cite{zhou2019accelerate}   & 1.78 & 0.56$\times$ & 71.50                \\ 
				RRBP\cite{zhou2019accelerate}  & 1.86 & 0.55$\times$ & 73.00                \\ 	
				GAL-0.1\cite{Lin2019Towards}   & 2.33 & 0.43$\times$ & -/71.95              \\ 
				GAL-0.5\cite{Lin2019Towards}   & 1.58 & 0.61$\times$ & -/69.88              \\ 
				CoGD-0.5 &  2.67    & 0.35$\times$ &  75.15/75.62         \\ 
				\hline\hline
		\end{tabular}}
		\label{imagenet_res50}
	\end{table}	

	For MobileNetV2, the pruning results are summarized in Tab.~\ref{cifar10_res}.  
	Compared to the pre-trained network, CoGD-$0.5$ achieves a \({\rm{0.45}} \times \) FLOPs reduction with a $0.18$\% accuracy drop. The results indicate that  CoGD is easily employed on efficient networks with depth-wise separable convolution, which is worth exploring in practical applications.

	\subsubsection{ImageNet}
	For ILSVRC12 ImageNet, we test our CoGD based on  ResNet-50. We train the network with a batch size of $256$ for $60$ epochs. The initial learning rate is set to $0.01$ and  scaled by $0.1$ over $15$ epochs. Other hyperparameters follow the settings used on CIFAR-10. The fine-tuning process follows the setting on CIFAR-10 with the initial learning rate $0.00001$.
	
   Tab.~\ref{imagenet_res50} shows that CoGD achieves state-of-the-art performance on the ILSVRC12 ImageNet. For ResNet-50, CoGD-$0.5$ further shows a  \({\rm{0.35}} \times \) FLOPs reduction while achieving only a $0.62$\% drop in accuracy.
	
	\subsubsection{Ablation Study}	
	We use ResNet-18 on CIFAR-10 for an ablation study to evaluate the effectiveness of our method. 
	
	\textbf{Effect on CoGD:} We train the pruned network with and without CoGD by using  the same  parameters. As shown in Tab.~\ref{ablation_study}, we obtain an error rate of $4.70$\% and a \({\rm{0.51}} \times \) FLOPs reduction 
	with CoGD, compared to the error rate is $5.19$\% and a \({\rm{0.32}} \times \)  FLOPs reduction without CoGD, validating the effectiveness of CoGD.
	
	\begin{figure}[!t]
		\centering
		\includegraphics[width=0.7\linewidth]{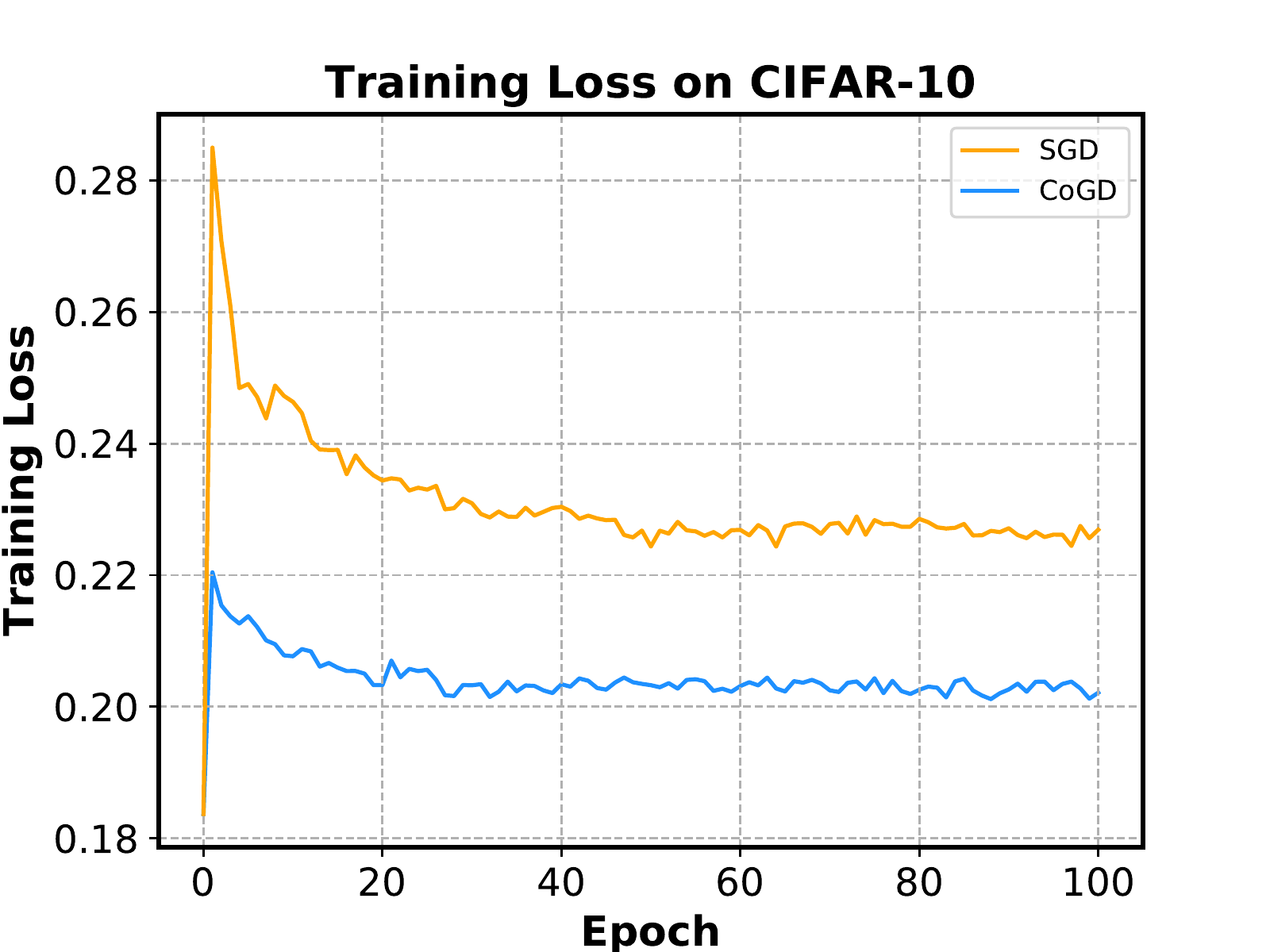}
		\caption{Comparison of training loss on CIFAR-10 with CoGD and SGD. }
		\label{convergence}
	\end{figure}		

	\textbf{Synchronous convergence:}
	In Fig.~\ref{convergence}, the training curve shows that the convergence of CoGD is similar to that of GAL with SGD-based optimization within an epoch, especially for the last epochs when converging in a similar speed. We theoretically derive CoGD within the gradient descent framework, which provides a theoretical foundation for the convergence, which is validated by the experiments.	As a summary, the main differences between SGD and CoGD are twofold. First, we change the initial point for each epoch. Second, we explore the coupling relationship between the hidden factors to improve a bilinear model within the gradient descent framework. Such differences do not change the convergence of CoGD compared with the SGD method.

\begin{figure}[htbp]
		\centering  
		\subfloat{
			\centering
			\label{synchronous}
			\includegraphics[width=0.7\linewidth]{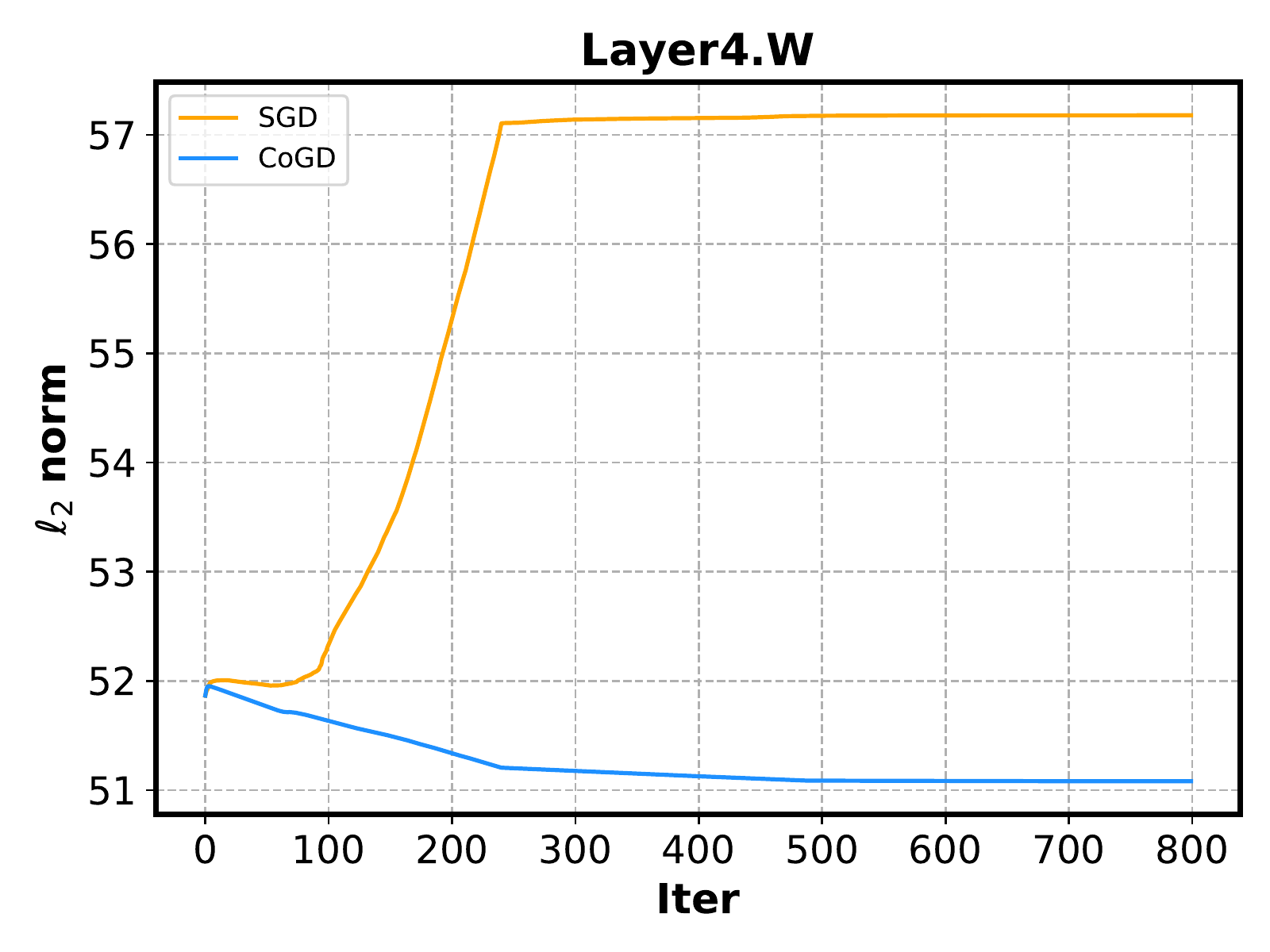}}
		\subfloat{
			\centering
			\label{asynchronous}
			\includegraphics[width=0.7\linewidth]{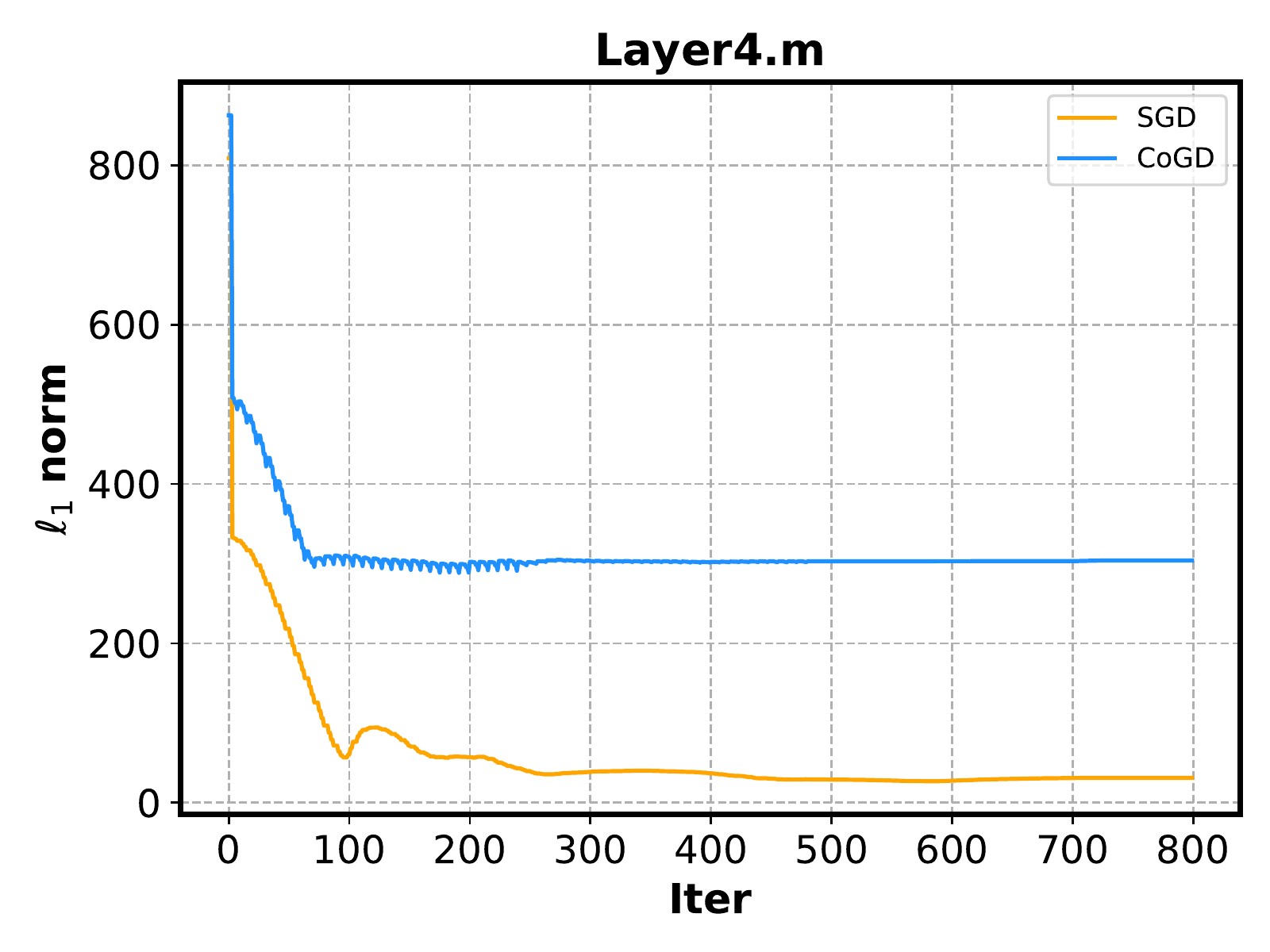}}
		\caption{Convergence comparison of the variables in the 4th convolutional layer when pruning CNNs. The curves are obtained using SGD and CoGD-$0.5$ on CIFAR-10. With CoGD, the two variables converge synchronously while avoiding either variable gets stuck in a local convergence (local minimum of the objective function), which validates that CoGD can avoid vanishing gradient for the coupled variables.}
		\label{fig:convergence}
	\end{figure}

	In Fig.~\ref{fig:convergence}, we show the convergence in a synchronous manner of the  $4$th layer’s variables when pruning CNNs. For better visualization, the learning rate of $\mathbf{m}$ is enlarged by 100$x$. On the curves, we observe that the two variables converge synchronously, and that neither variable  gets stuck into a local minimum.  This validates that CoGD avoids vanishing gradient for the coupled variables.
	
	\begin{table}[H]
		\centering
		\caption{Pruning results on CIFAR-10 with CoGD or SGD. M means million (\({10^6}\)).}
		\scalebox{0.9}{
			\begin{tabular}{c c c}
				\hline\hline
				Optimizer &Accuracy (\%) &FLOPs / Baseline (M)\\ \hline
				SGD &94.81  &376.12 / 555.43    \\		
				CoGD &95.30 &274.74 / 555.43    \\ \hline\hline
		\end{tabular}}
		\label{ablation_study}
	\end{table}

	\subsection{CNN Training}	
    Similar to network pruning, we have further evaluated CoGD algorithm for CNN model training on CIFAR-10 and ILSVRC12 ImageNet datasets. Specifically, we use ResNet-18 as the backbone CNN to test our algorithm. The network stages are 64-128-256-512. The learning rate is optimized by a cosine updating schedule with an initial learning rate $0.1$. The algorithm iterates $200$ epochs. The weight decay and momentum are respectively set to $0.0001$ and $0.9$. The model is trained on $2$ GPUs (Titan V) with a mini-batch size of $128$. We follow the similar augmentation strategy in \cite{He2016DeepNew} and add the cutout method for training. When training the model, horizontal flipping and $32 \times 32$ crop are used as data augmentation. The cutout size is set to $16$. Similar to CNNs pruning, $a$ is set to $0.95$ to compute  $\alpha_{\gamma}$ and $\alpha_{W}$. To improve the efficiency, we directly backtrack the corresponding weights. For ILSVRC12 ImageNet, the initial learning rate is set to $0.01$, and the total epochs are $120$. 

    With ResNet-18 backbone, we simply replace the SGD algorithm with the proposed CoGD for model training. In Tab.~\ref{cogd_cnn}, it can be seen that the performance is improved by $1.25$\% (70.75\% vs. 69.50\%) on the large-scale ImageNet dataset. In addition, the improvement is also observed on CIFAR-10. We report the results for different kernels, which show that the performance are relatively stable for $k = 1$ and $k = 2$. These results validate the effectiveness and generality of the proposed CoGD algorithm.

	\begin{table}[!ht]
		\small
		\caption{Results for CNN training on CIFAR-10 and ImageNet.}
		\centering
		\scalebox{0.83}{
			\begin{tabular}{c c c c c}
				\hline\hline
				\multicolumn{1}{c}{\multirow{2}{*}{Models}}& \multicolumn{2}{c}{Accuracy(\%)}  \\ 
				& \multicolumn{1}{c}{CIFAR-10}  &\multicolumn{1}{c}{ImageNet}\\\hline 
				ResNet-18 (SGD)\cite{He2016DeepNew} & 95.31 & 69.50 \\ 
				ResNet-18 (CoGD with K=1) & 95.80 & 70.30 \\
				ResNet-18 (CoGD with K=2) & 96.10 & 70.75 \\ 
				\hline\hline					
		\end{tabular}}
		\label{cogd_cnn}
	\end{table}

	\section{Conclusion}
	In this paper, we proposed a new learning algorithm, termed cogradient descent (CoGD), for the challenging yet important bilinear optimization problems with sparsity constraints. The proposed CoGD algorithm was applied on typical bilinear problems including convolutional sparse coding, network pruning, and CNN model training to solve important tasks including image reconstruction, image inpainting, and deep network optimization. Consistent and significant improvements demonstrated that CoGD outperforms previous gradient-based optimization algorithms. We further provided the solid derivation for CoGD, building an solid framework for leveraging a kernelized projection function to reveal the interaction of the variables in terms of convergence efficiency. Both the CoGD algorithm and the kernelized projection function provide a fresh insight to the fundamental gradient-based optimization problems.  
	\\



\acks{We are thankful for the comments from  Prof. Rongrong Ji  for their deep and insightful discussion. We would also appreciate the editors and anonymous reviewers for their constructive comments. }










\vskip 0.2in

\input{Cogradient.bbl}
\end{document}

%% file: algorithm/algorithm_CSC.tex
	\begin{algorithm}[tb]
	\caption{CoGD for CSC.}
	\label{alg_csc}
	\begin{algorithmic}[1]
		\REQUIRE The  training  dataset; sparsity factor \( \lambda\); hyperparameters such as penalty parameters, threshold $\alpha_{\mathbf{a}}$, $\alpha_{\mathbf{x}}$.
		\ENSURE The filters $\mathbf{a}$ and the sparse feature maps $\mathbf{x}$.
		\STATE Initialize $\mathbf{a}^0, \mathbf{x}^0$ and others
		\REPEAT
		\STATE Use $f_3({\mathbf{a}_k})$ in Eq.~\ref{eq:proximal_csc} to calculate kernel norm 
		\STATE 
		Use Eq.~\ref{bt_csc} to calculate $\mathbf{x}$
		\FORALL {$l = 1$ to $L$ epochs}
		\STATE Kernel Update:\\
		\(\mathbf{a}^l\) \( \gets \mathop {\arg \min }\limits_{\mathbf{a}}f_1(\mathbf{a}\mathbf{x})+\sum_{k=1}^K f_3(\mathbf{a}_k)\) using ADMM with proximal operators  
		\STATE Code Update:\\ 
		\(\mathbf{x}^l\) \( \gets  \mathop  {\arg \min}\limits_{\mathbf{x}}f_1(\mathbf{a}\mathbf{x})+\sum_{k=1}^K f_2(\mathbf{x}_k)\)  using ADMM with proximal operators  
		\ENDFOR
		
		\UNTIL{Loss convergence.}
	\end{algorithmic}
\end{algorithm}

%% file: algorithm/algorithm_pruning.tex
	\begin{algorithm}[tb]
	\caption{CoGD for  Pruning CNNs in Bilinear Modeling.}
	\label{alg_gal}
	\begin{algorithmic}[1]
		\REQUIRE
		The training dataset; the pre-trained network with weights $W_B$;
		sparsity factor \( \lambda\);
		hyperparameters such as learning rate, weight decay, threshold $\alpha_W$, $\alpha_{\mathbf{m}}$.
		\ENSURE
		The pruning network.
		\STATE Initialize \( W_G = W_B\), \(\mathbf{m} \sim N(0,1)\);
		\REPEAT
		\STATE Use $R(W_G)$ in Eq.~\ref{problem_pruning} to obtain the norms
		\STATE Use Eq.~\ref{bt_pruning} to calculate the new soft mask $\hat{\mathbf{m}}$			
		\FORALL {$l = 1$ to $L$ epochs}
		
		\FORALL {$i $ steps}
		\STATE Fix $W_G$  and update $W_D$ using Eq. \ref{optimization}
		\ENDFOR
		\FORALL {$j $ steps}
		\STATE Fix $W_D$  and update $W_G$  using Eq. \ref{optimization}
		\ENDFOR
		\ENDFOR
		
		\UNTIL{Loss convergence.}
	\end{algorithmic}
\end{algorithm}

%% file: algorithm/cnntraining.tex
	\begin{algorithm}[tb]
	\caption{CoGD for the CNN model training.}
	\label{alg_cnn}
	\begin{algorithmic}[1]
		\REQUIRE
		The training dataset; 
		sparsity factor \( \lambda\);
		hyperparameters such as learning rate, weight decay, threshold $\alpha_W$, $\alpha_{\mathbf{m}}$.
		\ENSURE
		CNN Model.
		\STATE Initialize and define the loss function (Eq. \ref{problem_cnnlearing}) with sparsity constraint on the kernel $W$;
		\REPEAT
		\STATE Use  Eq.~\ref{eq_convbc1} to define the BN layer to get a bilinear model
		\FORALL {$l = 1$ to $L$ epochs}

		\STATE Calculate $\hat{G}$ based on Eq. \ref{gcapcnn}
		\STATE Update $W$ based on Eq. \ref{bt_cnnlearning} and Eq. \ref{bch2} using  $\alpha_W$, $\alpha_{\mathbf{m}}$
		\STATE Update other parameters based on the autograd package in the Pytorch deep learning framework

		\ENDFOR
		
		\UNTIL{Loss convergence.}
	\end{algorithmic}
\end{algorithm}